\documentclass{article} 
\usepackage{iclr2025_re-align_workshop,times}

\usepackage{amsmath,amsfonts,bm}

\def\eqref#1{equation~\ref{#1}}

\def\1{\bm{1}}

\DeclareMathAlphabet{\mathsfit}{\encodingdefault}{\sfdefault}{m}{sl}
\SetMathAlphabet{\mathsfit}{bold}{\encodingdefault}{\sfdefault}{bx}{n}

\usepackage{hyperref}
\usepackage{url}
\usepackage{subcaption}
\usepackage{mathtools}
\usepackage{enumitem}
\usepackage{geometry}
\usepackage{multicol}
\usepackage{wrapfig}

\usepackage{amsmath}

\usepackage{dsfont} 
\usepackage{siunitx} 
\usepackage{booktabs}
\usepackage{xcolor}
\definecolor{anchorgreen}{RGB}{168, 214, 117}
\definecolor{termblue}{RGB}{119, 175, 223}
\definecolor{termpink}{RGB}{189, 113, 201}

\title{Cat, Rat, Meow: On the Alignment of Language Model and Human Term-Similarity Judgments}

\iclrfinalcopy 

\begin{document}
{\centering
\author{Lorenz Linhardt\textsuperscript{1,2,*} \And Tom Neuh{\"a}user\textsuperscript{1,2}\And Lenka T\v{e}tkov\'a\textsuperscript{3}\And Oliver Eberle\textsuperscript{1,2}}    
\maketitle
}

\renewcommand{\thefootnote}{\arabic{footnote}}
\footnotetext[1]{Machine Learning Group, Technische Universit\"at Berlin, Berlin, 10623, Germany}
\footnotetext[2]{BIFOLD - Berlin Institute for the Foundations of Learning and Data, Berlin, 10623, Germany}
\footnotetext[3]{Section for Cognitive Systems, DTU Compute, Technical University of Denmark, Kongens Lyngby, 2800, Denmark}
\renewcommand{\thefootnote}{\fnsymbol{footnote}}
\footnotetext[1]{Correspondence to: l.linhardt@tu-berlin.de} 

\begin{abstract}
Small and mid-sized generative language models have gained increasing attention. Their size and availability make them amenable to be analyzed at a behavioral as well as a representational level, allowing investigations of how these levels interact.
We evaluate 32 publicly available language models for their representational and behavioral alignment with human similarity judgments on a word triplet task. This provides a novel evaluation setting to probe semantic associations in language beyond common pairwise comparisons.
We find that (1) even the representations of small language models can achieve human-level alignment, (2) instruction-tuned model variants can exhibit substantially increased agreement, (3) the pattern of alignment across layers is highly model dependent, and (4) alignment based on models' behavioral responses is highly dependent on model size, matching their representational alignment only for the largest evaluated models. 
\end{abstract}

\section{Introduction}

Large language models (LLMs) have recently seen rapid progress, leading to the creation of numerous benchmarks, with great emphasis being placed on \textit{behavioral} evaluations (e.g.~\citep{srivastava2023beyond, Liu2023, liang2023holistic}). The demand for computational efficiency, accessibility, and privacy has driven the development of smaller language models~\citep{lu2024small}, which was made possible by advances in model distillation~\citep{gu2024minillm}, quantization~\citep{lin2024awq}, and pruning~\citep{wang2024comprehensive}.
Such models, which, unlike their larger counterparts, are often made publicly available, offer an opportunity to investigate the \textit{representations} underlying language model behavior. 
To understand language model representations and uncover relevant conceptual directions, methods such as representational similarity~\citep{kriegeskorte2008representational, kornblith2019similarity, klabunde2024resi}, probing and sparse autoencoders~\citep{bricken2023sae, cunningham2023sae}, manifold analysis~\citep{manfifold2020}, feature attribution approaches~\citep{eberle2020building, kauffmann2022clustering}, and function vectors~\citep{todd2024function} have been explored.
Studies comparing human signals to model predictions have revealed that models, even those trained on unrelated tasks like self-supervised prediction, show some representational alignment with human data of visual~\citep{doi:10.1073/pnas.1403112111, 8578166, Conwell2024} and language~\citep{abdou-etal-2021-language, pmlr-v235-huh24a, goldstein2024alignment} processing. However, significant differences in robustness, generalization, and alignment (e.g.~\citep{lapuschkin2019unmasking, geirhos2020shortcut, momennejad2023evaluating, muttenthaler2024improving}) between humans and deep models still persist, highlighting the need for a deeper understanding of these discrepancies.

In this work, we take a step towards understanding the structure of the internal representation spaces of language models. In particular, we use a triplet task from cognitive science that probes which words are considered more similar than others, and analyze the agreement of human similarity judgments with language model responses. We evaluate both representation similarities across multiple layers of recent models, as well as the models' behavioral (i.e. generative) responses. We seek to answer the following questions:
\begin{itemize}[itemsep=0pt,topsep=0pt]
\item[(Q1)] What is the general level of human alignment regarding similarity judgments, and how is it affected by instruction tuning and model size?
\item[(Q2)] How does this alignment change across layers? Can it be localized at a particular layer?
\item[(Q3)] Does representational alignment correspond to behavioral (generative) alignment?
\end{itemize}

We find that (1) the representation spaces of even small models are remarkably aligned with human similarity judgments and model size does not appear to be the main factor determining differences in alignment -- this stands in contrast to results on vision models, where alignment with human similarity judgments remains limited~\citep{Muttenthaler2023alignment}, (2) the pattern of representational alignment with human similarity judgments across the layers differs between models, (3) representations of instruction-tuned models are generally more aligned than their pretrained counterparts, (4) behavioral model evaluations show a clear correlation of model size and alignment -- the level of representational alignment is only reached by the largest models considered.

\section{Related Work}
\textbf{Textual semantic similarity} tasks have widely been used in cognitive science~\citep{tversky1977features, nosofsky1986attention, Hebart2020} and natural language processing~\citep{agirre-etal-2009-study, camacho-collados-etal-2017-semeval, 10.1145/3440755} to assess 
semantic similarity judgments in text. This required the collection of pair-wise similarity scores assigned by human raters, resulting in various datasets, e.g.~\citep{agirre-etal-2012-semeval, hill-etal-2015-simlex, Muennighoff2022MTEBMT}, typically containing few hundreds of samples. These datasets have so far mostly been used to evaluate or improve predictive capabilities of similarity models on pair-wise retrieval tasks~\citep{thakur2021beir, 2024-explaining-similarity,jiang-etal-2024-scaling}. 

\textbf{Triplet tasks}, instead of defining similarity via absolute pair-wise comparisons, rely on the relative evaluation of an anchor to potential targets~\citep{Hebart2020}. 
Earlier works have proposed the evaluation of vision model representations on triplet tasks~\citep{attarian2020, Muttenthaler2023alignment} to assess how well their representation spaces align with human similarity judgments (e.g. using the THINGS dataset~\citep{hebart2023THINGSdata}). This has led to techniques to align vision models post-hoc~\citep{muttenthaler2024improving, muttenthaler2024aligning}.
Recently, the triplet evaluation methodology was applied to language models~\citep{hrytsyna2024}. The authors translated THINGS images to text by captioning techniques, the choice of which had a significant impact on the results. In contrast, we employ the 3TT dataset~\citep{borghesani2023}, ensuring that the human and model responses are recorded for the same type of stimulus (text) and the effect of stimulus-specific context is minimized, e.g. no image background needs to be considered.

\section{Extracting Similarity Judgments From Humans and  Models}\label{sec:method}
\begin{figure}[thb]
    \centering
    \includegraphics[width=.9\textwidth]{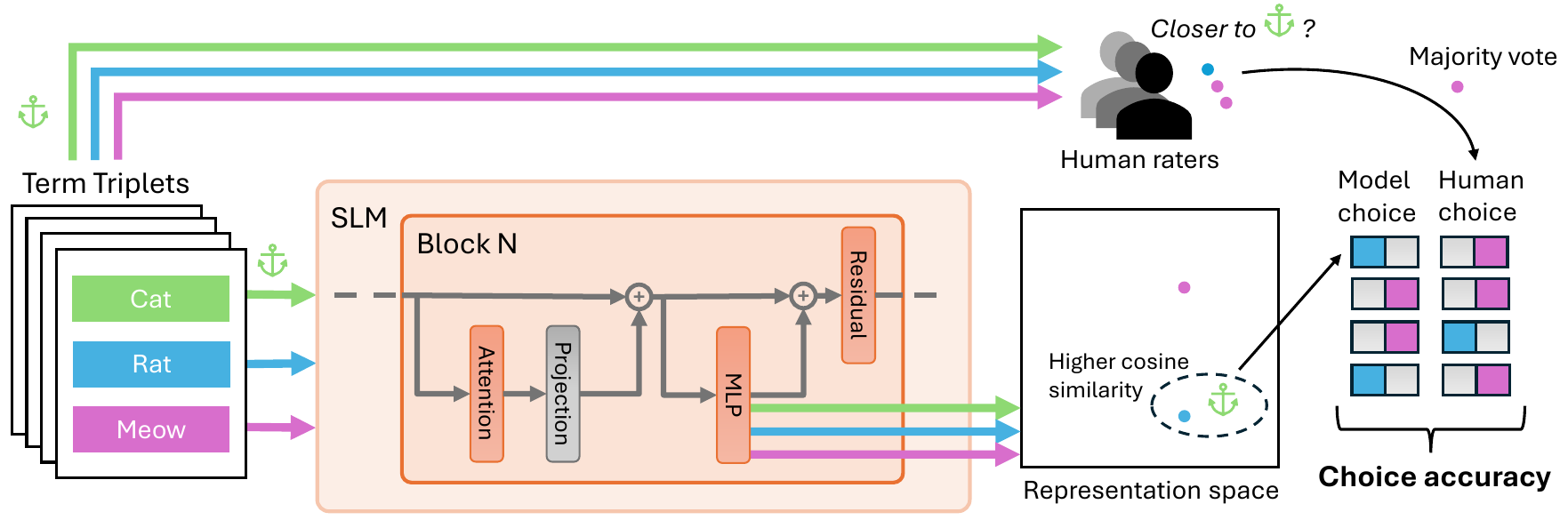}
    \caption{We assess the alignment of language model representations at different layers (attention block, MLP, residual stream) with the human similarity space via a triplet task: Human raters judge which of \textcolor{termblue}{two} \textcolor{termpink}{terms} is more similar to an \textcolor{anchorgreen}{anchor} term. The human choice is compared to the model choice, which is based on representational similarities to the anchor. The fraction of agreeing choices is the ``choice accuracy".
    }
    \label{fig:overview}
\end{figure}

\textbf{Human Similarity Judgments.}\hspace{.4em}
Different experimental designs exist to extract human similarity ratings of concepts. As asking for a numerical similarity score for two items suffers from mismatching scales across different human raters~\citep{Hebart2020}, triplet-task designs have emerged as an alternative (e.g.~\citep{fukuzawa1988, robilotto2004, Li2016, Hebart2020, borghesani2023}). In the design used in this work, given terms A, B, and an anchor C, raters are asked a variation of: \textit{Which of the terms, A or B, is closer in meaning to C?} The resulting choice is binary (scale-free) and allows to gauge \textit{relative} distances of concepts. 

We use the three-terms-task (3TT) dataset~\citep{borghesani2023}, containing \num{10107} term triplets sampled from \num{6433} unique words. Of these triplets \num{2555} have been labeled by a set of \num{1322} raters, with at least 17 judgments per triplet.
A majority response can be calculated for each triplet, as well as an \textit{agreement} score, which is the absolute difference of raters choosing the first and second target term, divided by the total number of judgments. As our primary objective is the comparison of model representations with human judgments, we only use the subset of human-evaluated triplets. 
In the following, we consider the \textit{majority vote} of all human raters for any triplet as the ``human choice''. If both choices have an equal number of votes, we omit the triplet from our analysis, leaving $N=2539$ triplets.

\textbf{Model Response Extraction.}\hspace{.4em}
To extract and evaluate model responses based on representations, we follow the pipeline shown in Fig.~\ref{fig:overview}. By ``model choice,'' we denote the term which, when embedded at a given layer, has larger cosine similarity to the anchor term.
We consider the fraction of model choices that agree with the human choice as our measure of alignment with the human similarity space, which we name \textit{choice accuracy}.
To additionally extract behavioral model responses, we prompt the instruction-tuned models with an adapted version of the prompt used for the creation of the 3TT dataset~\citep{borghesani2023}.
Details on the extraction of representational and behavioral responses are summarized in Appx.~\ref{appx:method}.

\section{Evaluation of Language Models on the 3TT Dataset} 

We evaluate a set of 32 language models\footnote{obtained from  \url{www.huggingface.co}} from 6 model families (Gemma 2~\citep{gemma2}, LLama 3~\citep{llama3}, Minitron~\citep{minitron}, OpenELM~\citep{openelm}, Phi~\citep{phi15, phi2, phi3}, and Qwen 2.5~\citep{qwen25}) on the 3TT dataset. 17 of the models are only \textit{pretrained}, and 15 are \textit{instruction tuned} after pretraining.

\subsection{(Q1) How Well Aligned are the Representations of Language Models? }\label{sec:ex:unc}

\begin{wraptable}{L}{5.5cm}
\vspace{-5pt}
    \centering
    \setlength{\tabcolsep}{4pt}
\scriptsize
\begin{tabular}{lccccc}
\toprule
Model & Pretr. & I. T. & Behav. & Invalid\\
\midrule
Gemma2-2B & 0.77 & \textbf{0.79} & 0.70 & 0.01 \\
Gemma2-9B & 0.77 & 0.82 & \underline{\textbf{0.83}} & 0.03 \\
Llama-3.1-8B & 0.81 & \textbf{0.82} & \textbf{0.82} & 0.00 \\
Llama-3.2-1B & 0.78 & \textbf{0.80} & 0.48 & 0.01\\
Llama-3.2-3B & 0.80 & \textbf{0.81} & 0.69 & 0.00\\
Minitron-4B & 0.59 & \textbf{0.78} & 0.67 & 0.07 \\
Minitron-8B & \textbf{0.61} & - & - & - \\
OpenELM-270M & 0.\textbf{79} & \textbf{0.79} & 0.00 & 1.00 \\
OpenELM-450M & \textbf{0.82} & 0.81 & 0.00 & 1.00 \\
OpenELM-1.1B & \textbf{0.81} & 0.80 & 0.00 & 1.00 \\
OpenELM-3B & \textbf{0.77} & \textbf{0.77} & 0.00 & 1.00 \\
Phi-1.5 & \textbf{0.79} & - & - & - \\
Phi-2 & \textbf{0.79} & - & - & - \\
Phi-3.5-mini & - & 0.77 & \textbf{0.78} & 0.03 \\
Qwen-2.5-0.5B & 0.66 & \textbf{0.80} & 0.44 & 0.04 \\
Qwen-2.5-1.5B & 0.67 & \textbf{0.78} & 0.57 & 0.01 \\
Qwen-2.5-3B & 0.67 & \textbf{0.79} & 0.70 & 0.07 \\
Qwen-2.5-7B & 0.66 & \textbf{0.79} & \textbf{0.79} & 0.02 \\
\midrule
Models (mean) & 0.79 & \textbf{0.82} & \textbf{0.82} & - \\
\midrule
\midrule
LSN & - & - & \textbf{0.74} & - \\
Humans (mean) & - & - & \textbf{0.82} & - \\
\bottomrule
\end{tabular}
    \caption{Maximum choice accuracy across all layers for (1) representations of pretrained, (2) representations of instruction-tuned (I.T.), and (3) behavior of instruction-tuned models. Bold numbers are the row-wise maximum, underlined is the overall maximum. The last column indicates the fraction of invalid model answers. The mean choice accuracy over human choices is ``Humans (mean)" and over all valid model choices is ``Models (mean)". A dash indicates inexistent models and evaluations.}
    \label{tab:max-ca}
\vspace{-20pt} 
\end{wraptable}

To assess whether language models produce similarity choices akin to humans', we investigate: (1) choice accuracy, which serves as a basic indicator of the alignment of representation spaces, and (2) whether the ratio of distances of the two choices to the anchor corresponds to the level of agreement between humans.

\textbf{Do language models capture human term similarities well?}\hspace{.4em} In Tab.~\ref{tab:max-ca}, we report the choice accuracy of the layer achieving the highest choice accuracy for each model.
To contextualize the results, we provide the average human alignment with the majority vote (fraction of human choices agreeing with the human majority vote) as well as the neuro-cognitive inspired model Lancaster Sensorimotor Norms~\citep{lynott2020lsn} (LSN), both of which can be derived from the 3TT dataset.
We find that (1) even the smallest models can be more accurate in predicting term similarities than LSN, which is a strong baseline~\citep{borghesani2023}, and (2) while there are variations across models, their choice accuracies are close to or even reach the average human choice accuracy.
We provide examples of triplets for which the model choice consistently (dis)agrees with the human majority in Appx.~\ref{appx:examples}. For example, for the anchor \textit{cat} and targets \textit{rat} and \textit{meow}, most humans choose \textit{meow} but all pretrained models pick \textit{rat} as the most similar term.

\textbf{Does model size matter?}\hspace{.4em}
Across all models, we find that a higher number of parameters neither consistently positively nor consistently negatively affects representational choice accuracy. Notably, the pretrained model with the highest choice accuracy is OpenELM with 450M parameters. We conclude that for models in the evaluated parameter range, a low number of parameters does not prevent learning representations aligned with human similarity judgments. 

\begin{wrapfigure}{l}{5.5cm}
\vspace{-5pt}
    \centering
        \includegraphics[width=\linewidth]{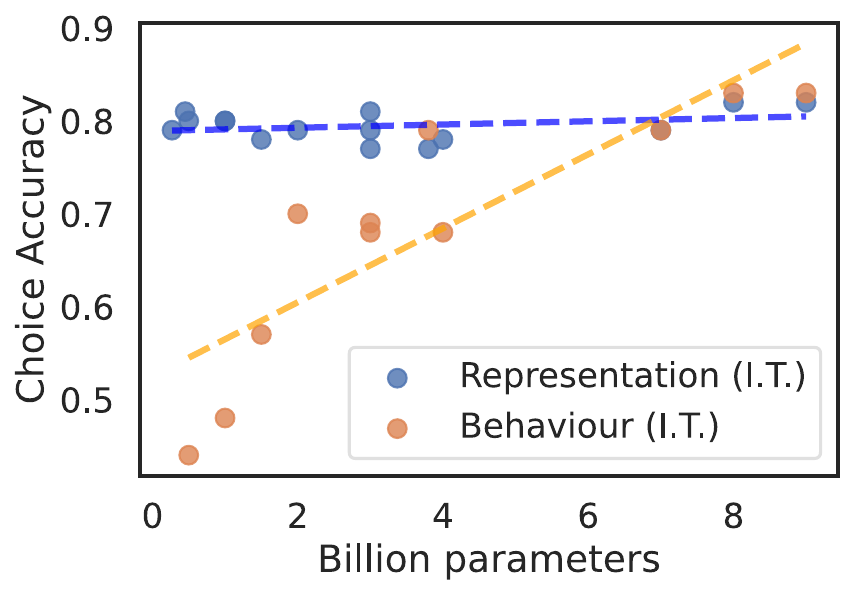}
    \caption{Representational and behavioral choice accuracy v.s. model size for instruction-tuned (I.T.) models. OpenELM models are excluded due to poor instruction following.}
    \label{unc:fig:ca-vs-params}
    \vspace{-23pt}
\end{wrapfigure}

\textbf{What is the impact of instruction tuning?}\hspace{.4em}
For Gemma 2, Minitron, and Qwen-2.5 models we observe a substantial increase in choice accuracy, with the latter two families showing an increase of 0.1 to 0.2 in choice accuracy. In these cases, instruction tuning appears to have aligned the internal representation spaces with human similarity judgments. For no model does instruction tuning have a significant negative effect on choice accuracy.

\textbf{Do relative similarities model human agreement?}\hspace{.4em}
To further investigate the correspondence of models' and humans' similarity judgments, we evaluate whether the relative representational similarity of the two choices to the anchor corresponds to human disagreement. For this purpose, we define a quantity $\gamma := \rho(a,1-c)$,
where $\rho$ is the Pearson correlation coefficient, $a \in [0,1]^N$ is a vector of human agreement scores per triplet (see Sec.\ref{sec:method}) and $c \in [0,1]^N$ is calculated as the \textit{distance ratio} of the smaller and the larger cosine distance of the targets to the anchor. Intuitively, the distance ratio quantifies how clearly the model prefers one choice over the other.

We find that while there is a strong correlation of choice accuracy and $\gamma$ ($r=.95$, $p<.001$ for instruction-tuned models and $r=.93$, $p<.001$ for pretrained models), the correlation of agreement and distance ratios is weak ($\gamma<0.4$). This reveals that the distance ratio of models' representations poorly models human disagreement and that this mismatch is unaffected by model size. We refer to Appx.~\ref{appx:calibration} for more detailed results.

\subsection{(Q2) How Does Alignment Vary Across Layers?}
\begin{figure}[ht]
    \centering
        \includegraphics[width=\linewidth]{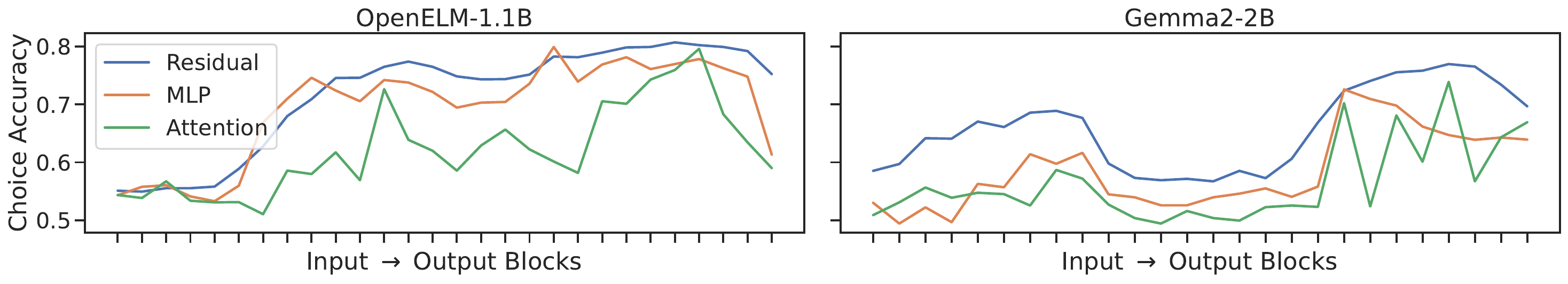}
    \vskip -0.1in
    \caption{Choice accuracy across layers for two pretrained models. (\textbf{Left}) in OpenELM-1.1B, choice accuracy rises nearly monotonically, (\textbf{right}) in Gemma2-2B, a bimodal pattern can be observed.}
    \label{layers:fig:2examples}
    \vskip -0.15in
\end{figure}

We observe that even though all models follow the same architectural structure (blocks of attention and MLP layers, operating on the residual stream), the pattern of choice accuracy levels across layers differ by model family. Fig.~\ref{layers:fig:2examples} shows one such example: whereas in OpenELM-1.1B, choice accuracy increases rather monotonically until the last third of the model, in Gemma 2-2B, we can see a bimodal choice-accuracy profile. Across all models, choice accuracy appears to increase up until later layers, indicating that the relative arrangement of terms in representation space changes significantly over the layers, even if no context is provided. Instruction tuning appears to modulate the progression of choice accuracy across layers, e.g. in Gemma 2 models, the observed bimodality is considerably flattened. We refer to Appx.~\ref{appx:ca_layers} for plots for the remaining models, as well as to Appx.~\ref{appx:deepseek} for additional results on reasoning models. Furthermore, we find that residual stream representations often achieve the models' highest choice accuracies and do not see as strong fluctuations in choice accuracy as attention or MLP layers (see Appx.~\ref{appx:ca_layers}). Cases where residual stream layers are superseded in choice accuracy usually see single attention layers achieving the maximum.

\subsection{(Q3) Does Representational Alignment Correlate with Behavioral Alignment?}
One of the core questions of the representational alignment community is to what extent one can translate between observations of representational structure and behavioral outcomes. While the 3TT dataset only provides behavioral outcomes for human participants, we can observe in models to what extent representational choice accuracy is correlated with behavioral choice accuracy.  

Unlike in the evaluation of representational alignment, it can be seen in Tab.~\ref{unc:fig:ca-vs-params} that behavioral alignment increases with model size.
This pattern is evident in models such as Qwen-2.5-7B, where both metrics are closely aligned. In contrast, smaller models (e.g., Qwen-2.5-0.5B) show poor behavioral alignment, which cannot be attributed only to failures in adhering to the expected output format, indicated by higher rates of invalid answers. OpenELM models are the exception, almost always failing to match the answer format. 

Overall, these results suggest that model scale plays a critical role in achieving behavioral alignment and opens the possibility that representational alignment forms an approximate upper bound on behavioral alignment. Furthermore, our results suggest that small language models may contain more knowledge than can be extracted from them in generative evaluations.

\section{Discussion and Conclusion}
In this work, we found that even small language models can show human-like agreement with the human majority choice on a term similarity task. This is remarkable since the task does not exactly specify what type of similarity (e.g. lexical, concept feature, concept associative~\citep{borghesani2023}) is to be used for making choices. We further find that alignment is often positively impacted by instruction tuning. Most studies so far have focused on behavioral alignment~\citep{10.1093/pnasnexus/pgae233, wang-etal-2023-self-instruct, chia-etal-2024-instructeval}, with recent evidence suggesting a positive impact of instruction tuning also on representational alignment~\citep{aw2024instructiontuning}. Interestingly, a models' behavioral alignment, unlike representational alignment, is dependent on model size. While the choice accuracy of language models is high, we found that the anchor-target similarity in representation space does not capture human disagreement well.

We believe that extending these analyses to more complex textual data and task-based evaluations can pave the way for the automatic assessment of language models' representational structure,  potentially uncovering spurious associations and discrepancies in concept alignment.
Furthermore, triplet-term data may be used, similar to work on vision models, to encourage human-like representation structures, making the model more robust and trustworthy. 

\section*{Acknowledgements}
The authors would like to thank the anonymous reviewers for their constructive feedback.
We acknowledge funding by the German Ministry for Education and Research (refs. 01IS18037A and 01IS18025A). This work was supported by the Novo Nordisk Foundation grant NNF22OC0076907 ”Cognitive spaces - Next generation explainability”.

\bibliographystyle{main}
\bibliography{main}

\clearpage
\appendix

\section{Details to Model Response Extraction}\label{appx:method}
In this section, we provide additional details on how model responses were extracted to calculate choice accuracy from both representational and behavioral responses.

\subsection{Representational responses}
    \begin{figure*}[htb!]
        \subfloat{%
            \includegraphics[width=.49\linewidth]{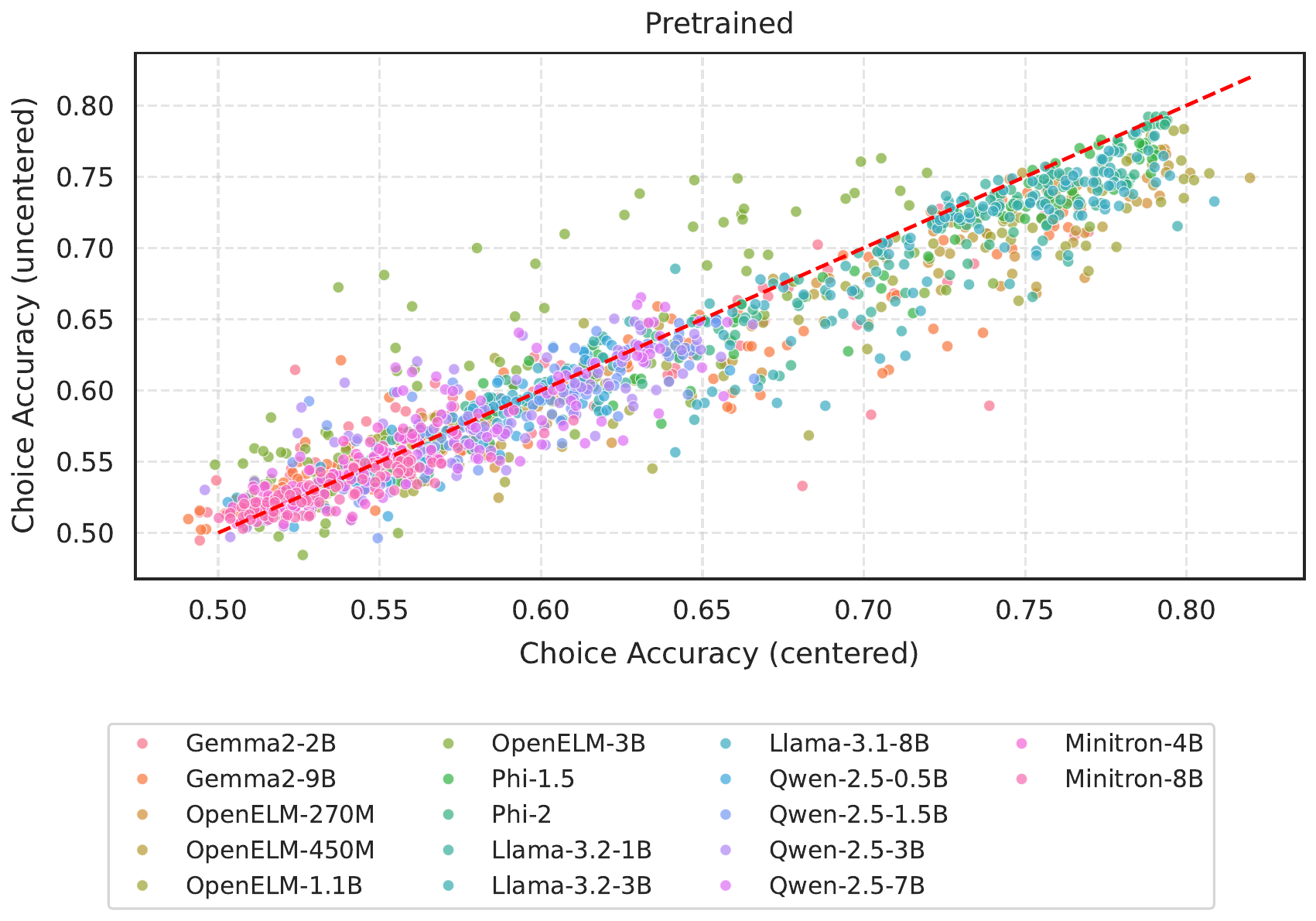}%
        }\hfill
        \subfloat{%
            \includegraphics[width=.49\linewidth]{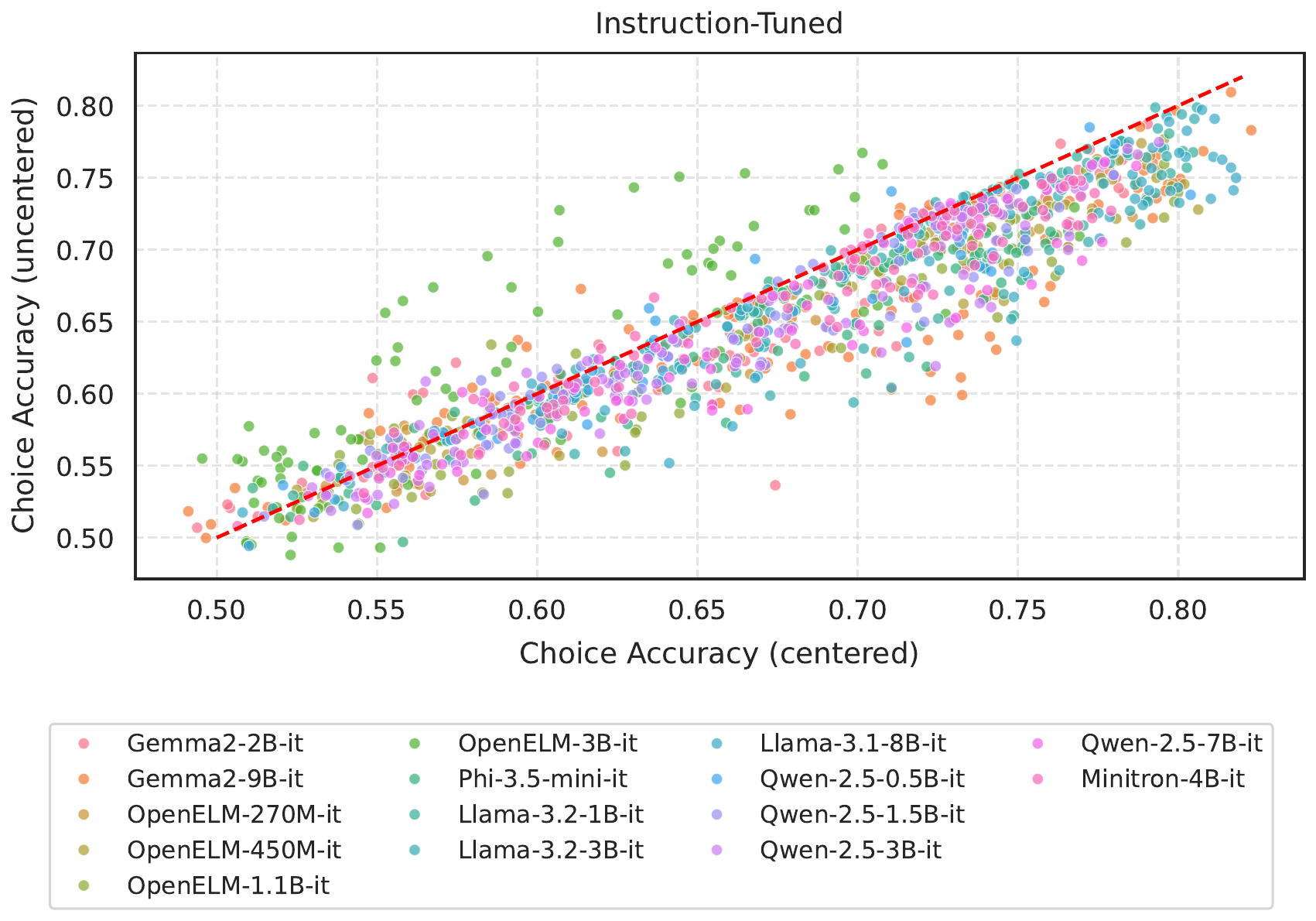}%
        }
        \caption{Choice-accuracy before and after representation centering for pretrained (\textbf{left}) and instruction tuned (\textbf{right}) models. In most cases, centering leads to higher choice accuracy (below the diagonal).}
        \label{appx:fig:centering}
    \end{figure*}

For extracting representations from pretrained (not instruction-tuned) models, we embed each term individually. Special \texttt{<bos>} tokens are prepended if required by the model. For each word, we record the representation after each sub-block (attention, MLP, residual stream). Instruction-tuned model variants are fed the individual terms in the \texttt{user} part of the corresponding chat template and empty \texttt{system} prompts. In both cases, the last token corresponding to the input term is recorded.

All representations used for calculating representational alignment are centered per layer (i.e. the mean over the layer-specific representations of all terms is subtracted).
As we use cosine similarity as a basis of the three-terms-task evaluation, moving the origin of the representation space can significantly impact the results. It can be seen in Fig.\ref{appx:fig:centering} that in most cases, centering leads to small improvements in choice accuracy.

\subsection{Behavioral responses}
To extract behavioral responses from language models for the three terms task, we prompt the instruction-tuned models with an adapted version of the instructions for human raters used for the creation of the 3TT dataset~\citep{borghesani2023}. The adapted prompt was designed to reduce the rate of invalid answers: \textit{``Which of the words A or B is closer in meaning with the word C? Answer with exactly one word: either A or B. Do not answer with C. Do not answer in a full sentence."} We post-process the models' responses by removing special characters and transforming them to lowercase. To compute the choice accuracy, we determine whether the post-processed response equals \textit{A} or \textit{B} or neither, which we count as invalid choice. To preclude potential bias introduced by the ordering of the presented choices, we randomize their order and report the average choice accuracy over 3 seeds.
In Appx.~\ref{appx:promt-robust}, we provide additional analyses on the robustness of the prompt concerning term order and instruction complexity. 

\clearpage
\section{Choice Accuracy Across Layers}\label{appx:ca_layers}

\begin{wrapfigure}{r}{0.4\textwidth}
    \centering
    \includegraphics[width=\linewidth]{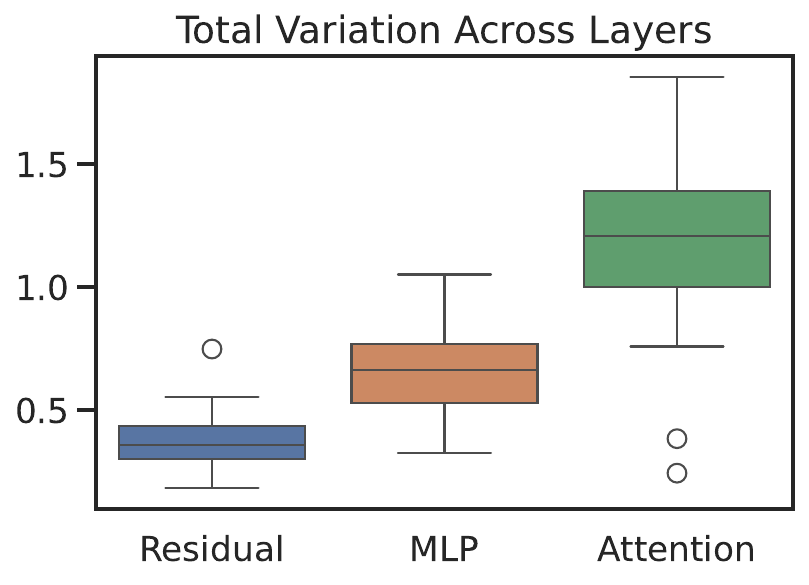}
    \caption{Total variation of choice accuracy over all layers of a particular type. Each box aggregates all pretrained models.}
    \label{appx:fig:tx_boxes}
\end{wrapfigure}

\begin{figure}[]
    \centering
    \begin{multicols}{2} 
        \includegraphics[width=\linewidth]{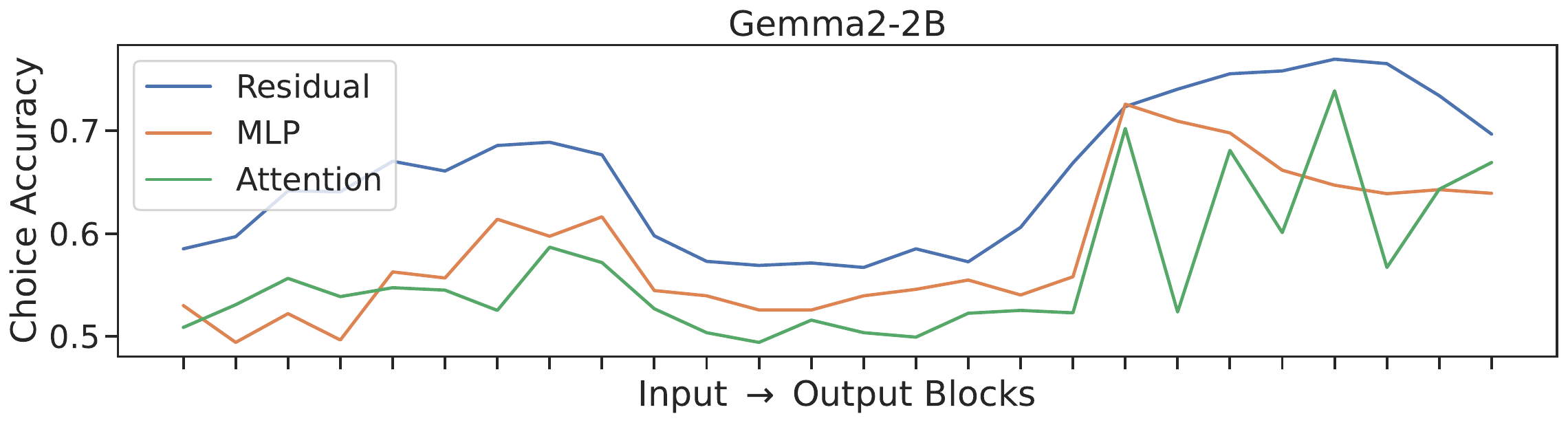}\par
        \includegraphics[width=\linewidth]{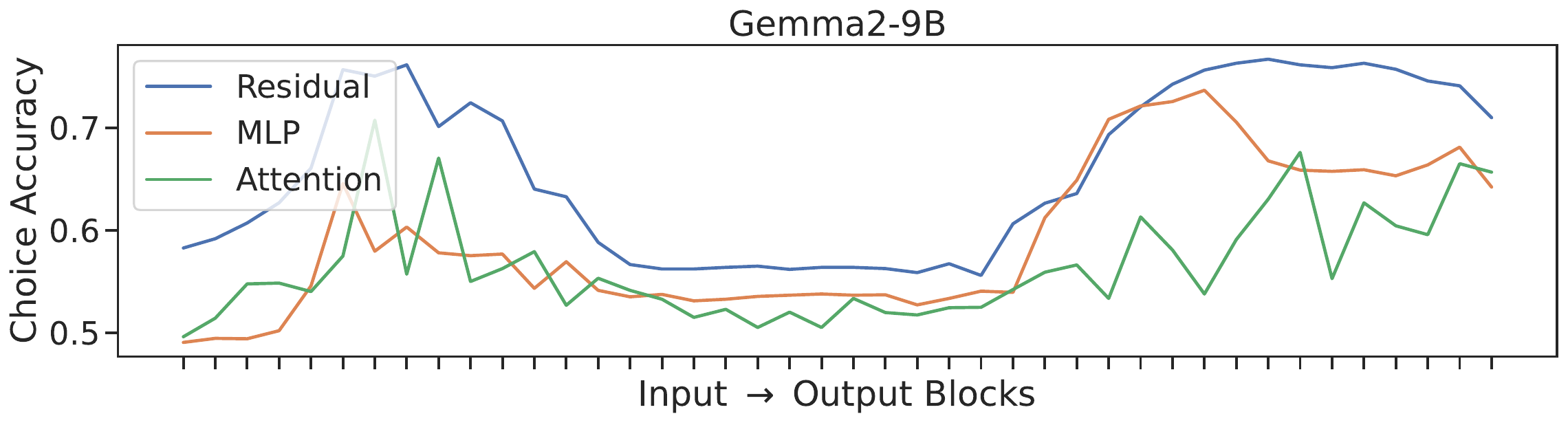}\par
        \includegraphics[width=\linewidth]{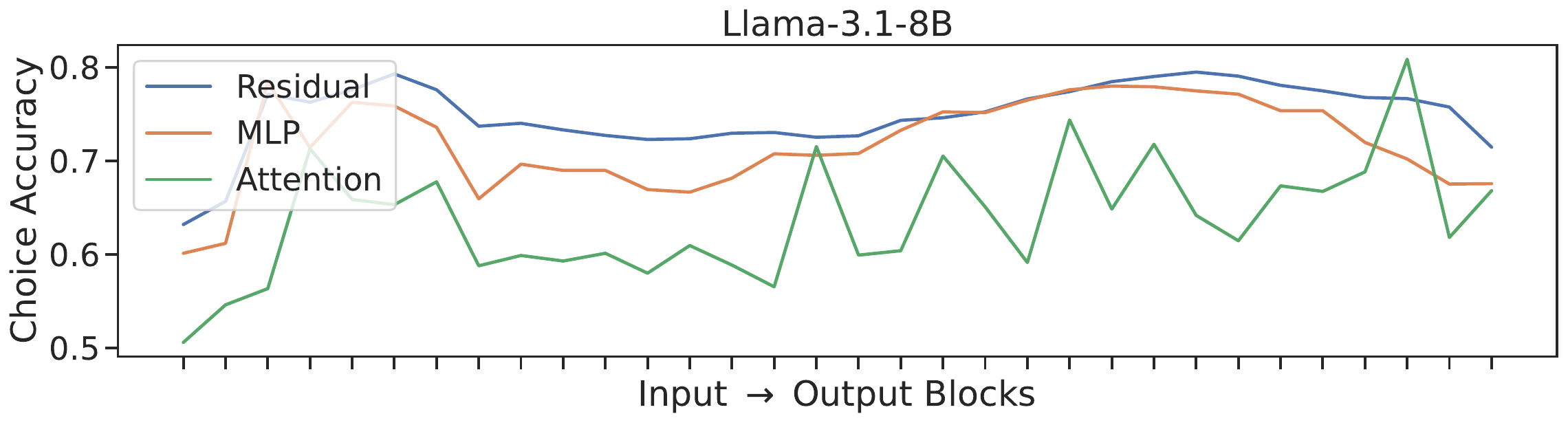}\par
        \includegraphics[width=\linewidth]{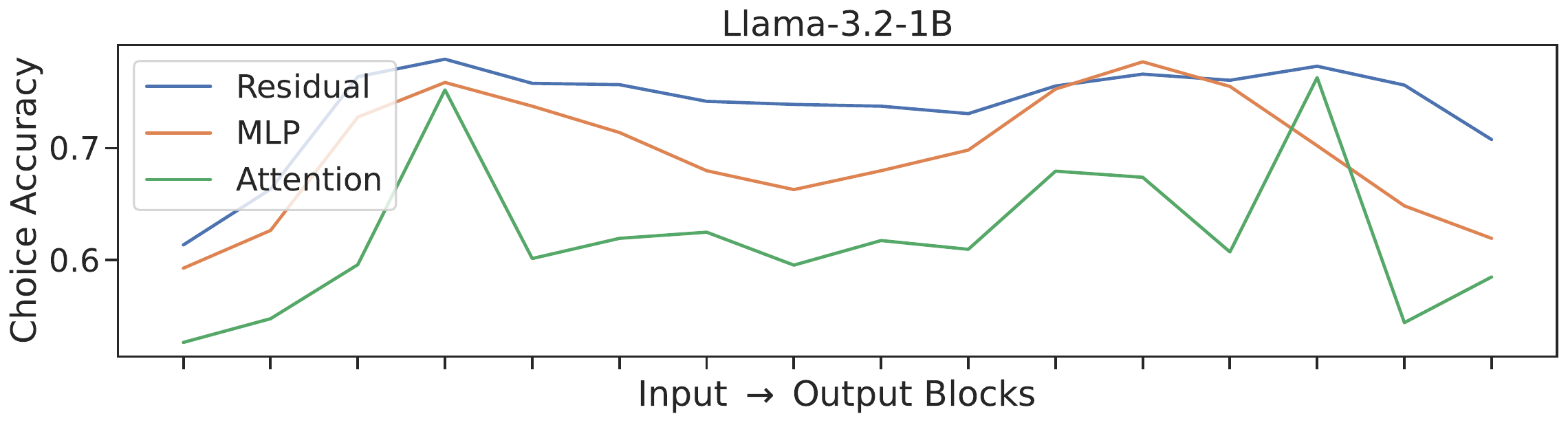}\par
        \includegraphics[width=\linewidth]{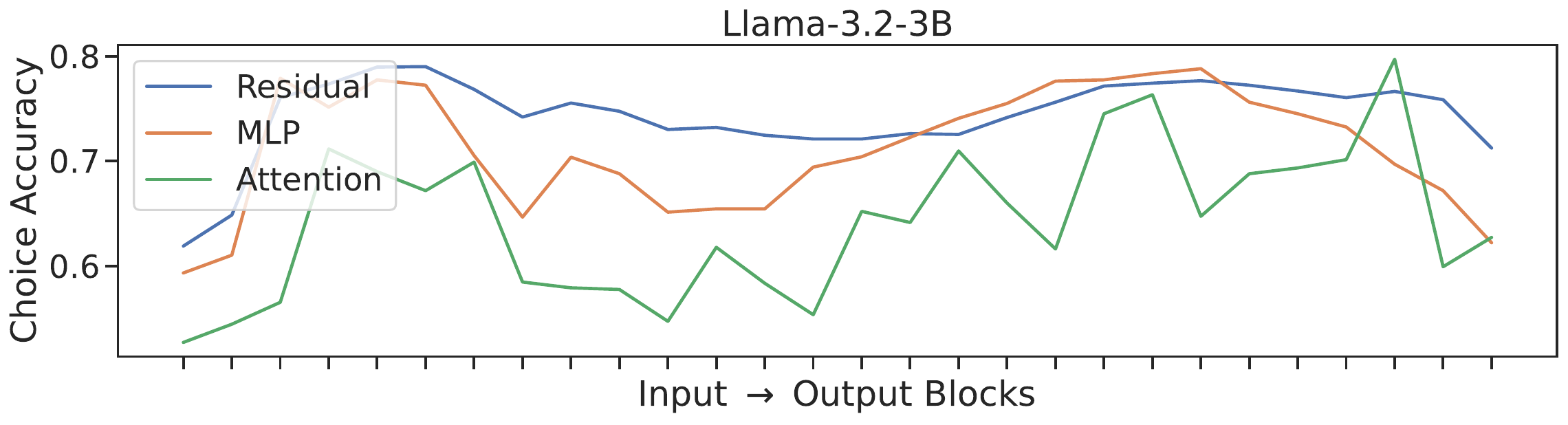}\par
        \includegraphics[width=\linewidth]{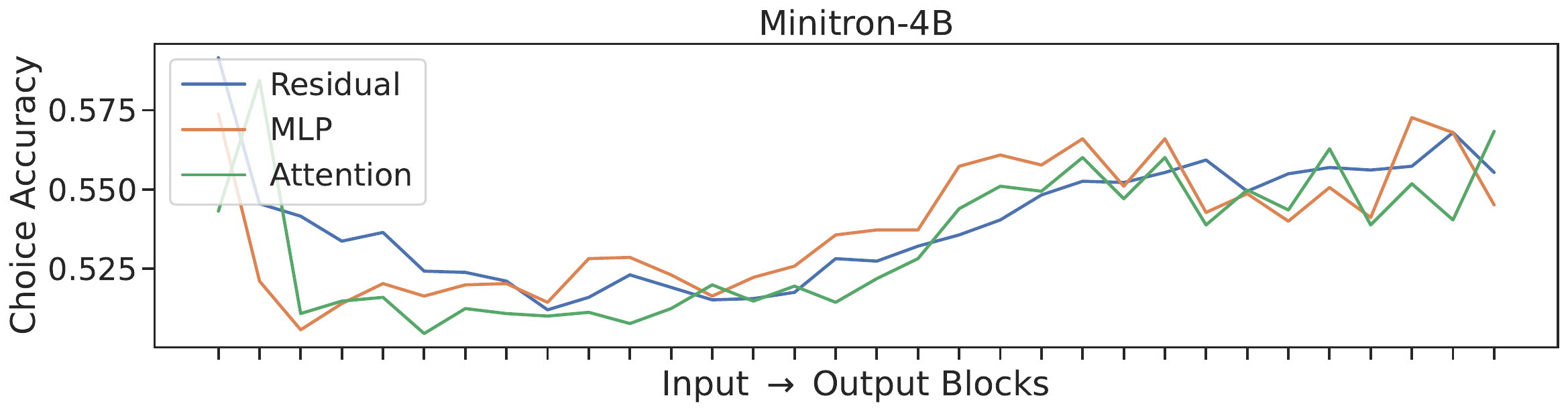}\par
        \includegraphics[width=\linewidth]{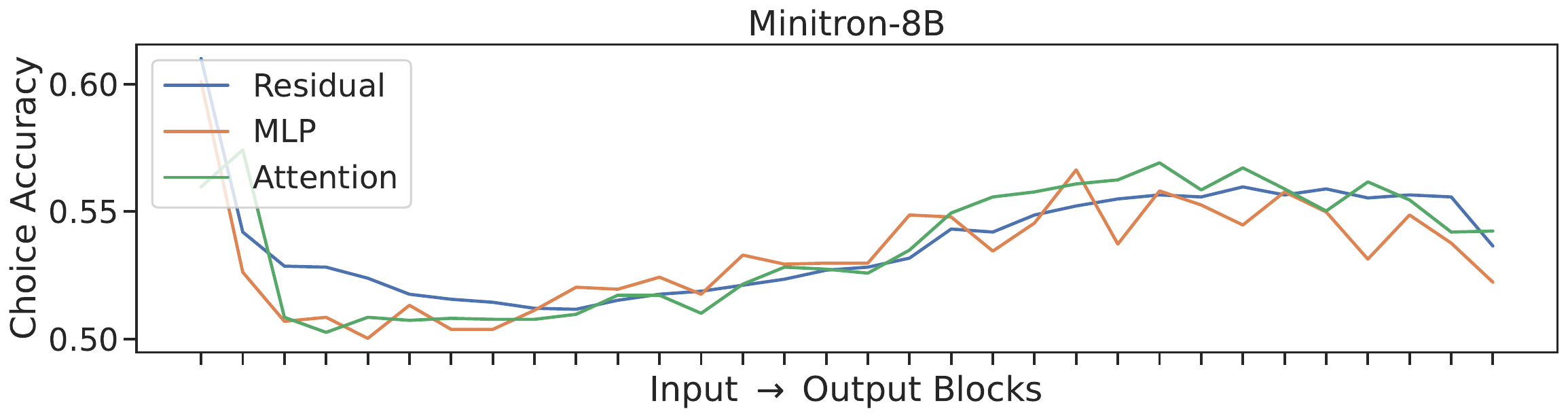}\par
        \includegraphics[width=\linewidth]{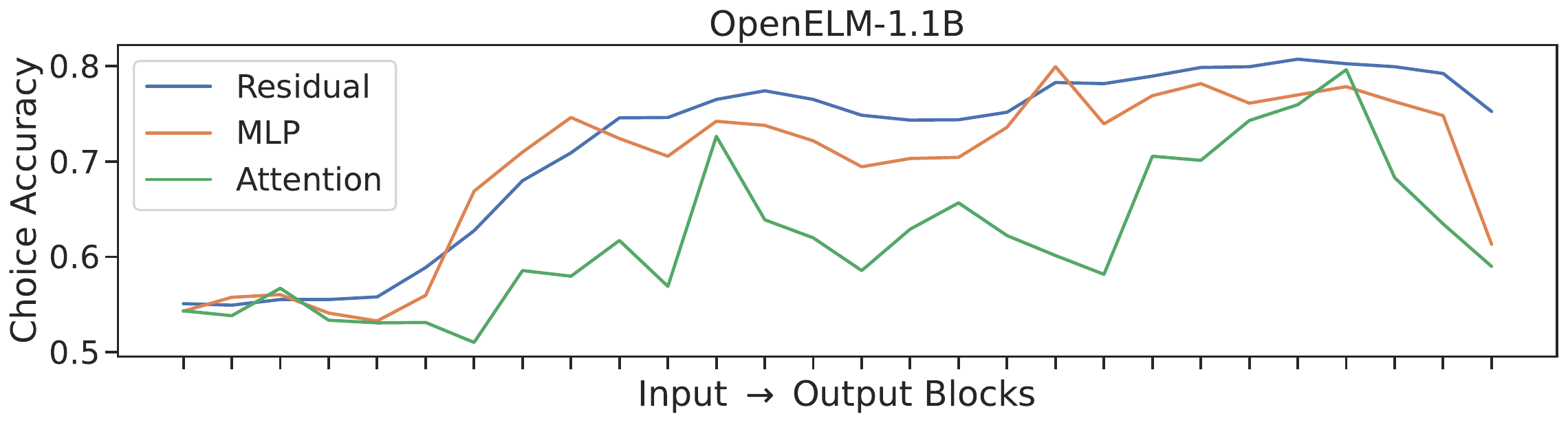}\par
        \includegraphics[width=\linewidth]{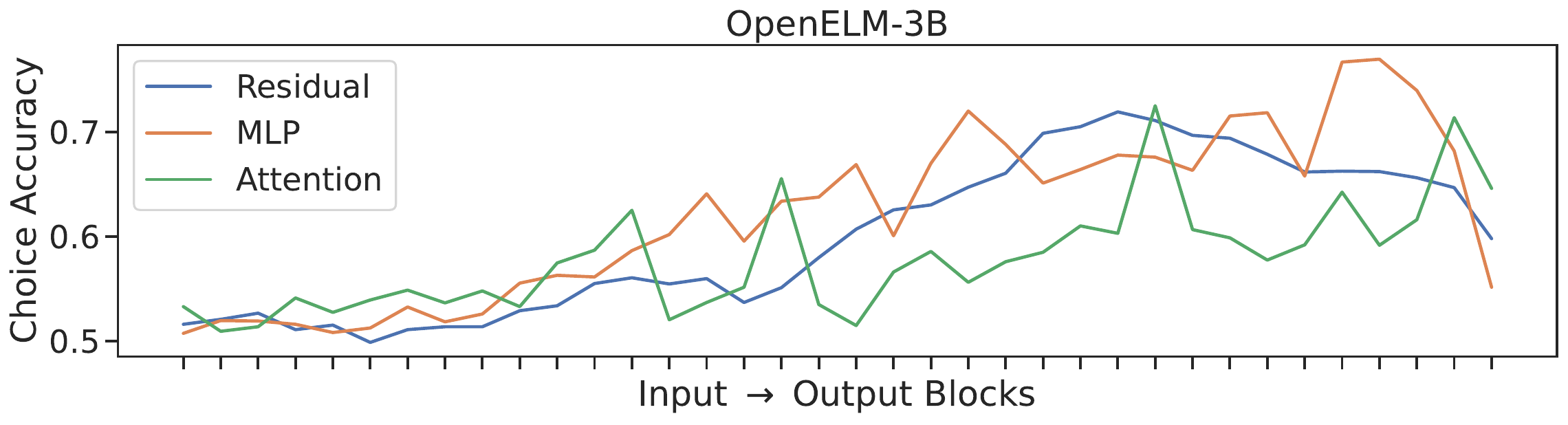}\par
        \includegraphics[width=\linewidth]{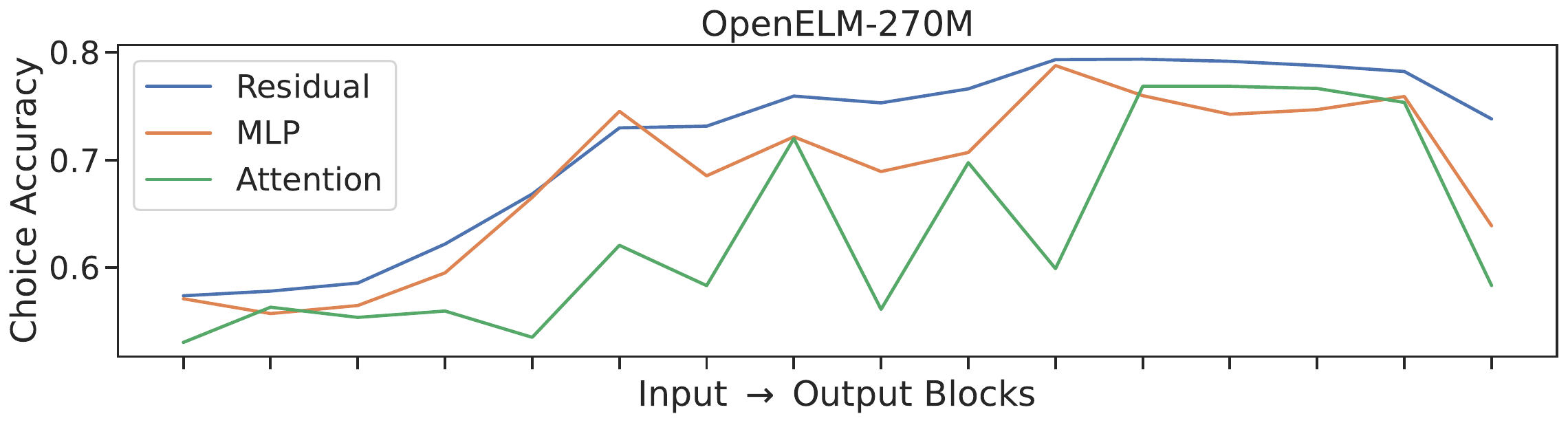}\par
        \includegraphics[width=\linewidth]{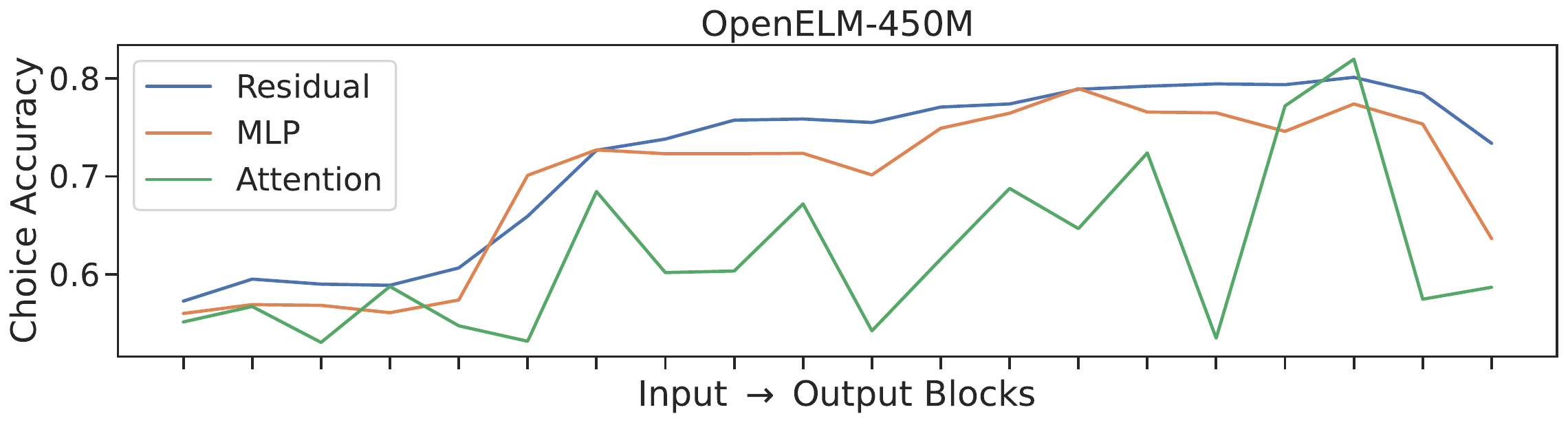}\par
        \includegraphics[width=\linewidth]{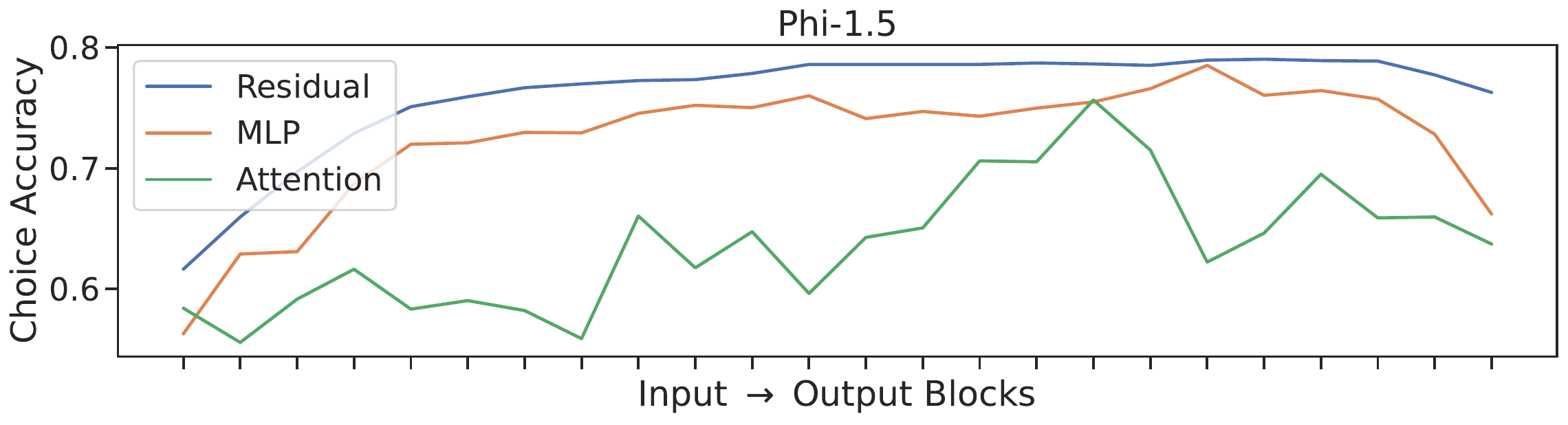}\par
        \includegraphics[width=\linewidth]{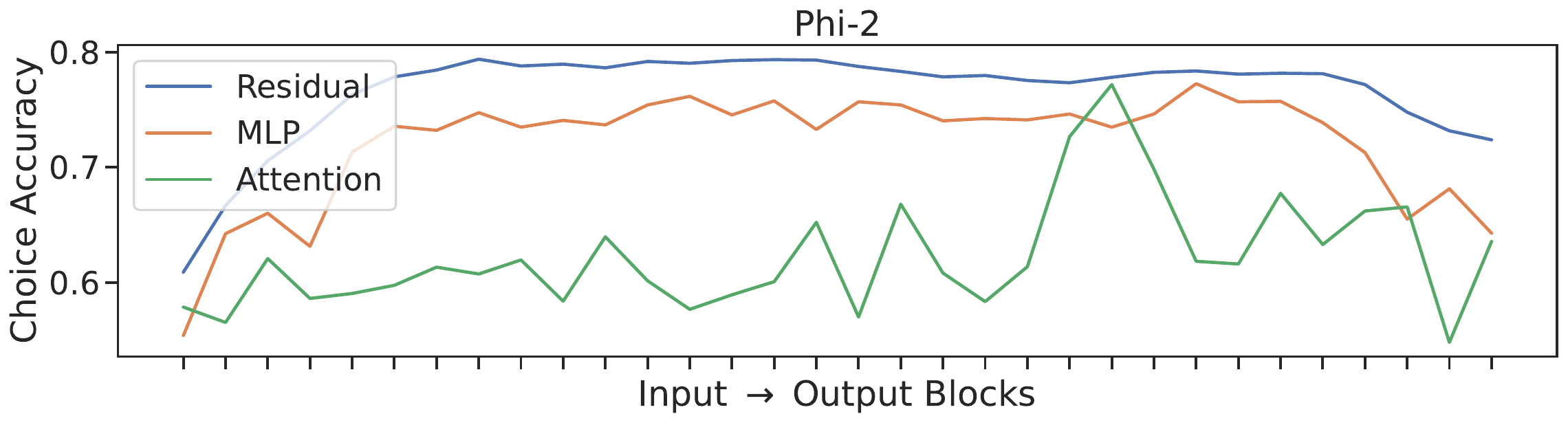}\par
        \includegraphics[width=\linewidth]{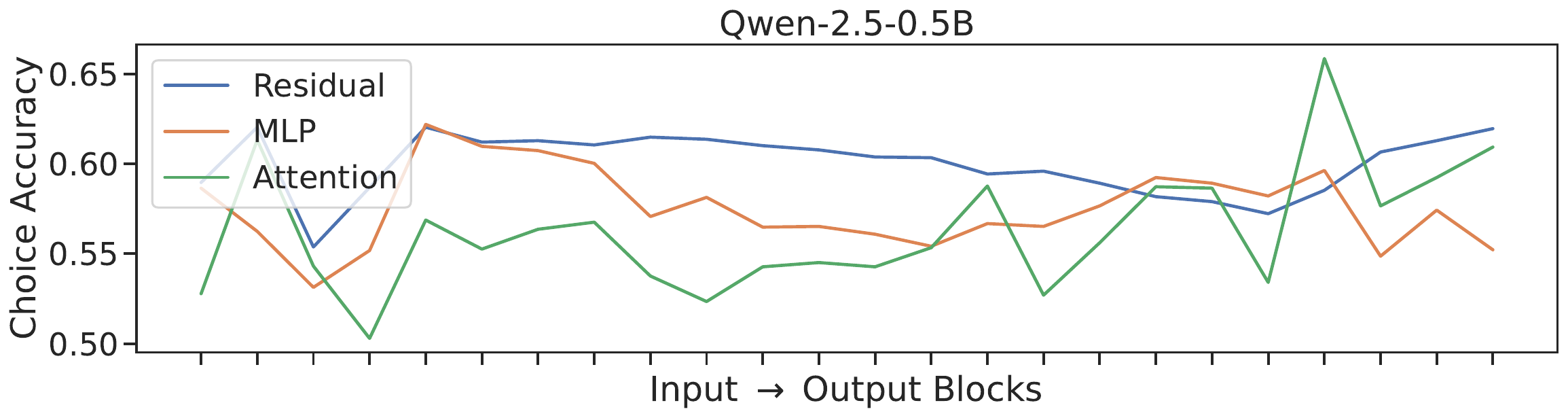}\par
        \includegraphics[width=\linewidth]{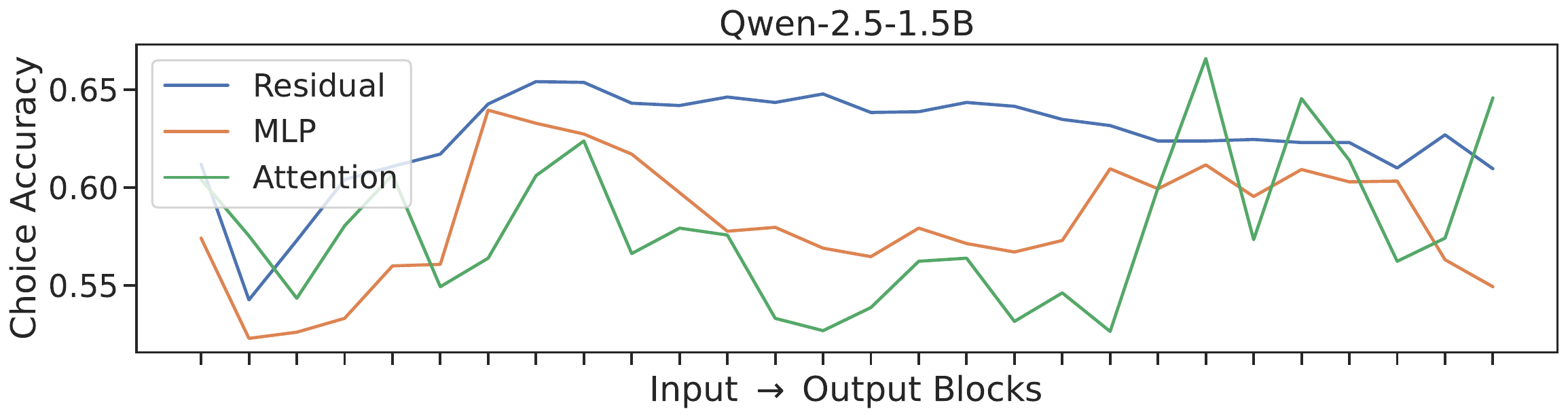}\par
        \includegraphics[width=\linewidth]{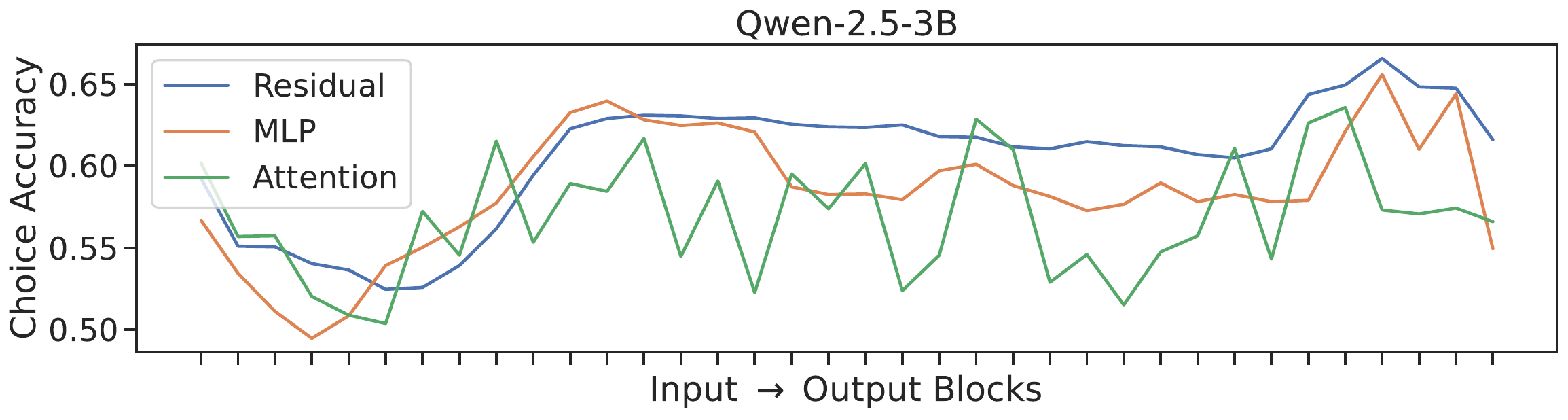}\par
        \includegraphics[width=\linewidth]{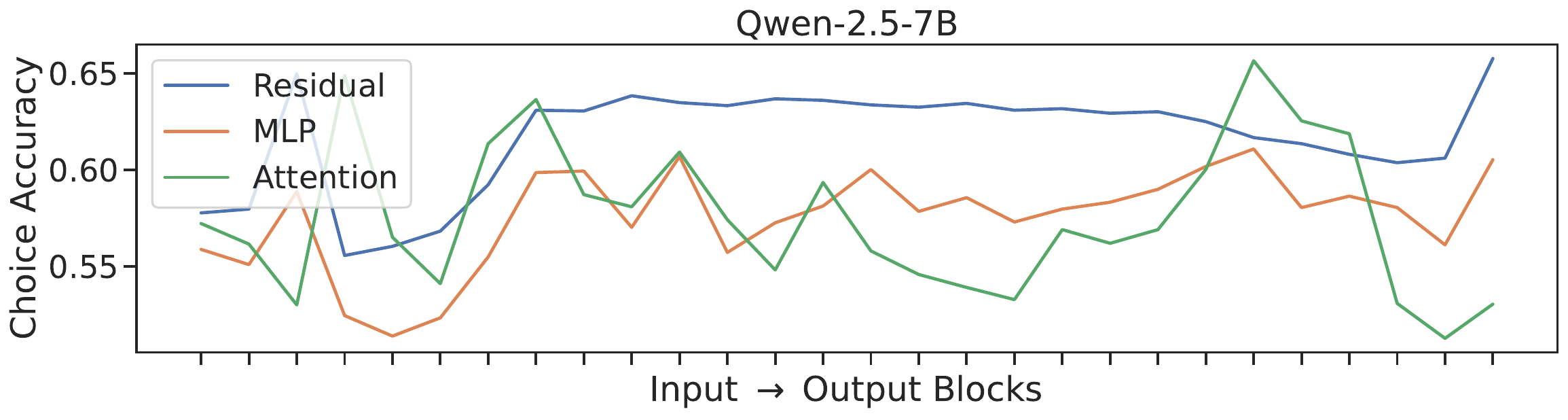}\par
    \end{multicols}
    \caption{Development of choice accuracy over layers for all pretrained models.}
    \label{appx:fig:ca_all_pt}
\end{figure}

\begin{figure}[]
    \centering
    \begin{multicols}{2} 
        \includegraphics[width=\linewidth]{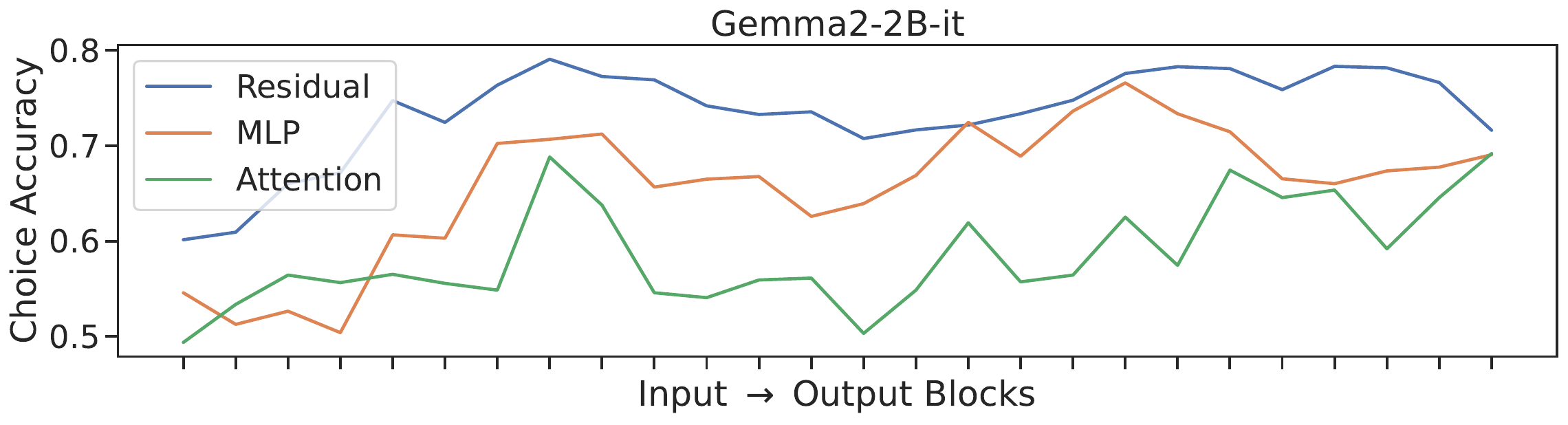}\par
        \includegraphics[width=\linewidth]{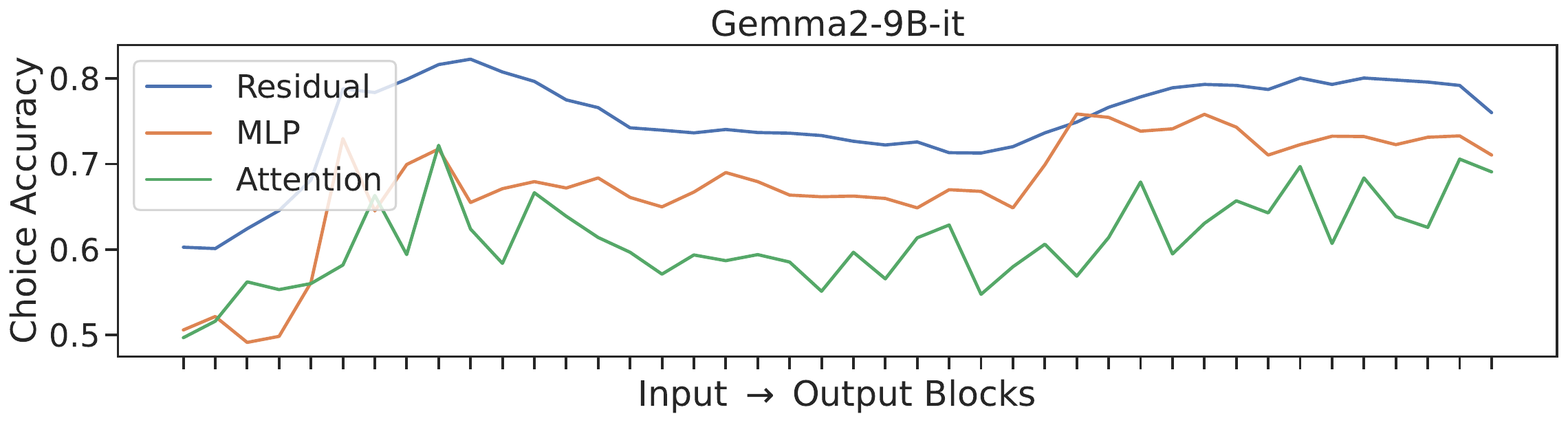}\par
        \includegraphics[width=\linewidth]{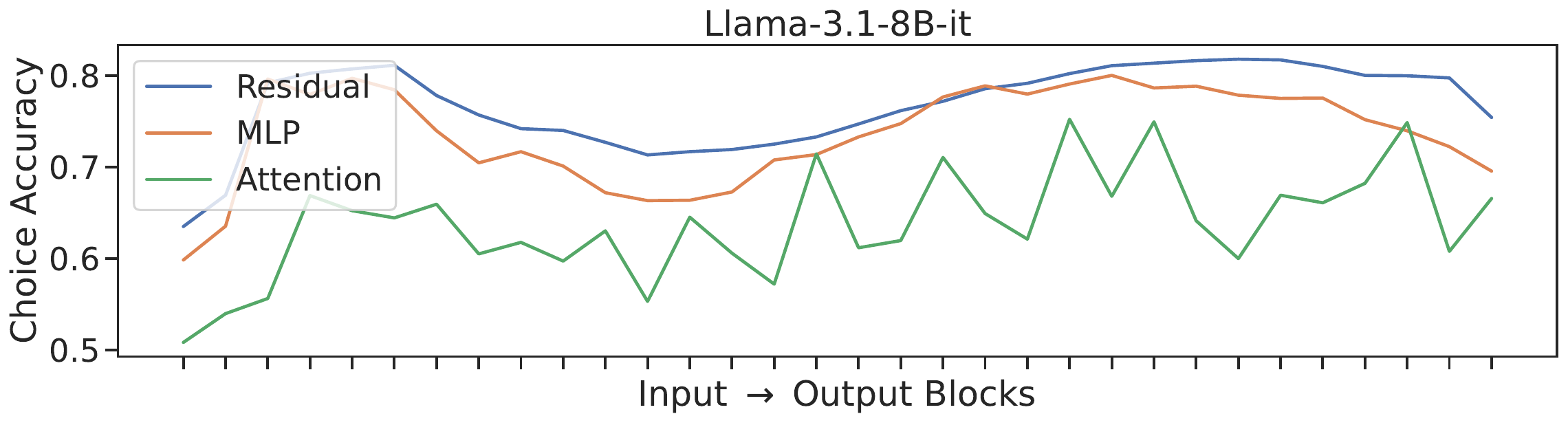}\par
        \includegraphics[width=\linewidth]{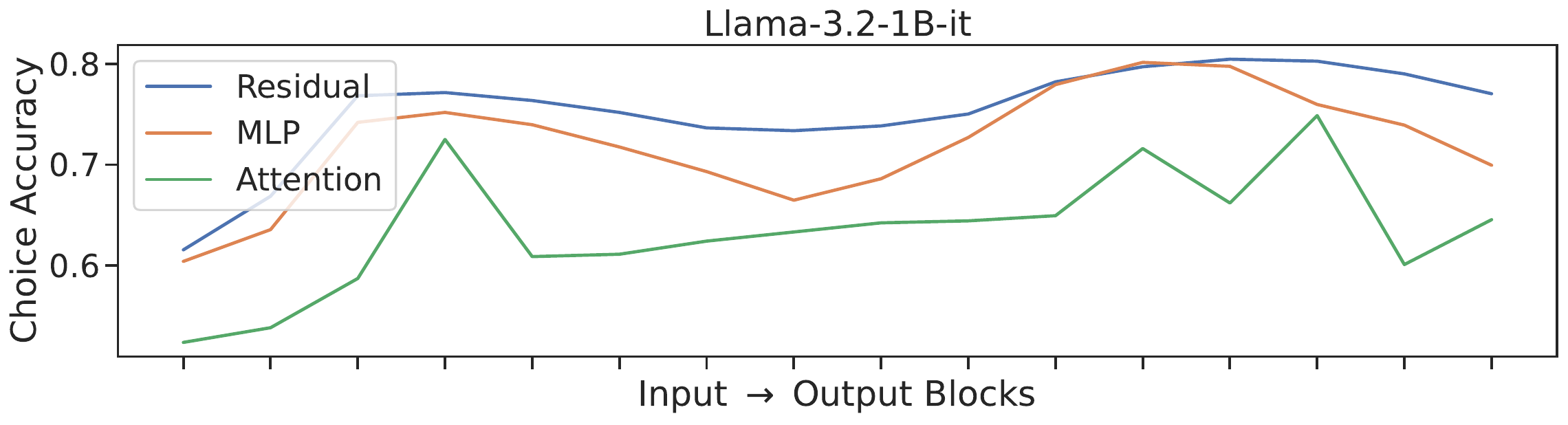}\par
        \includegraphics[width=\linewidth]{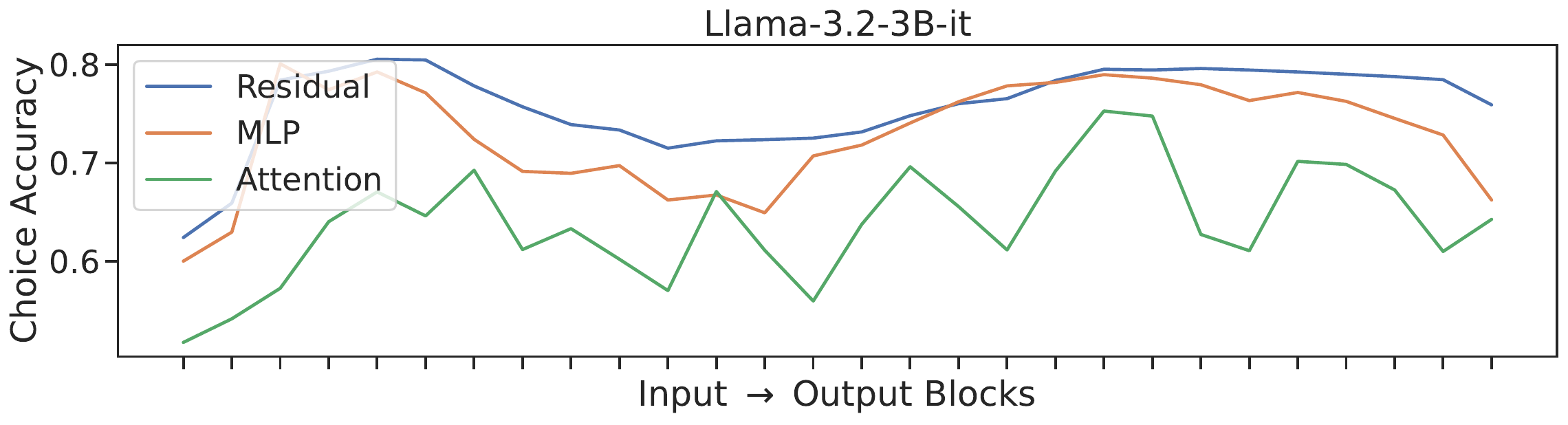}\par
        \includegraphics[width=\linewidth]{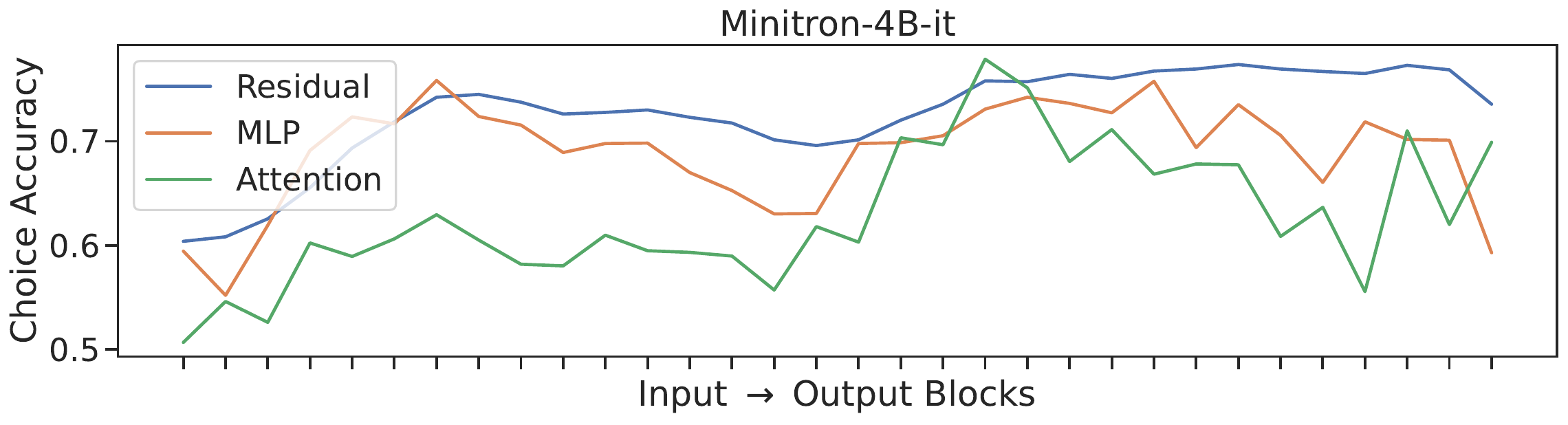}\par
        \includegraphics[width=\linewidth]{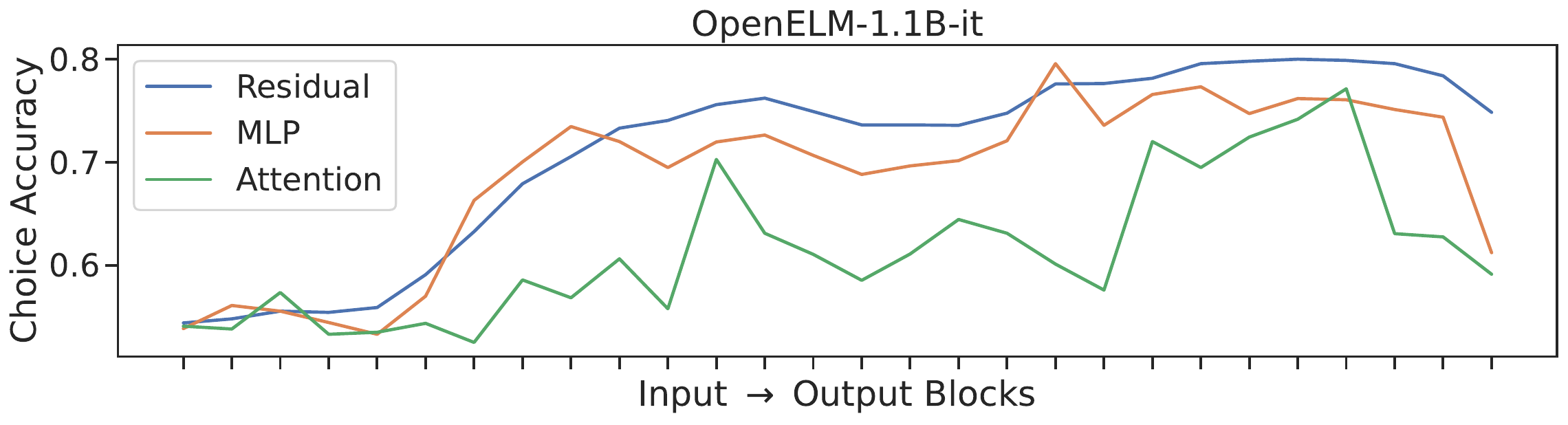}\par
        \includegraphics[width=\linewidth]{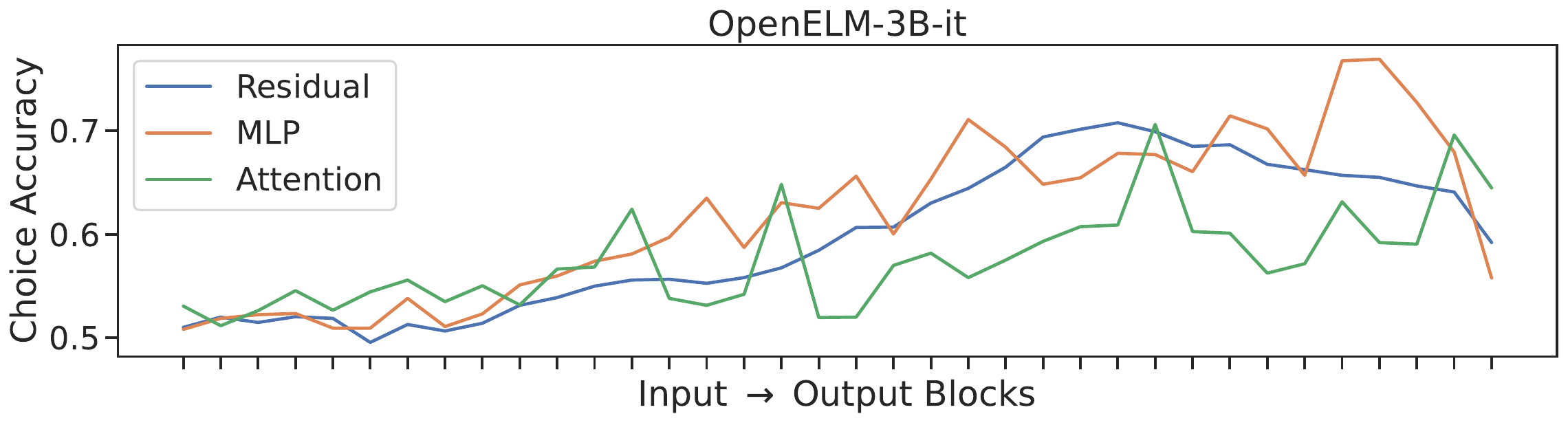}\par
        \includegraphics[width=\linewidth]{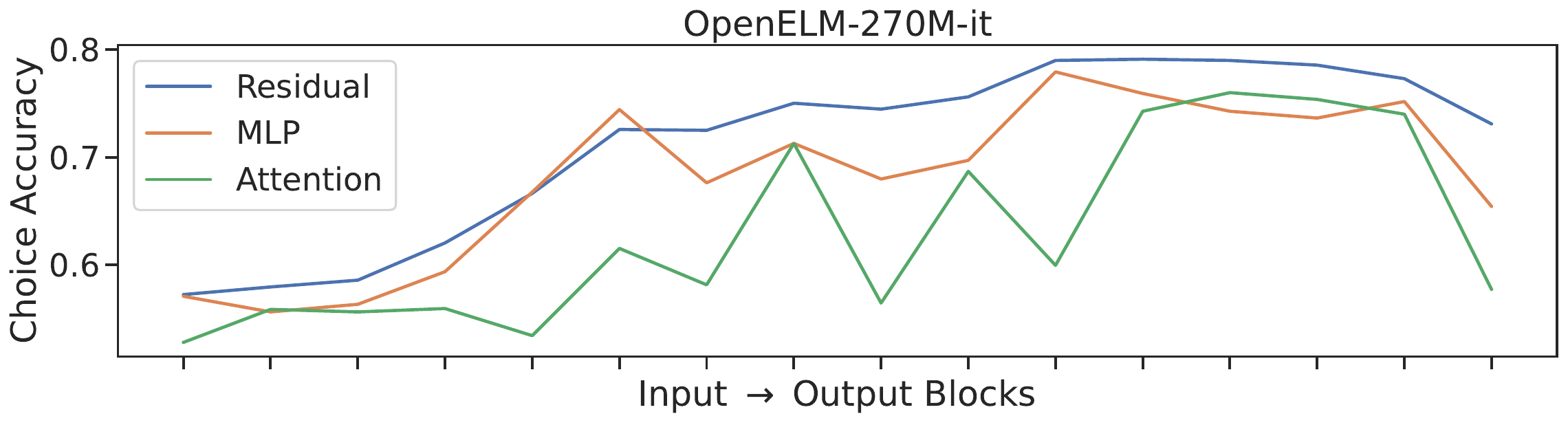}\par
        \includegraphics[width=\linewidth]{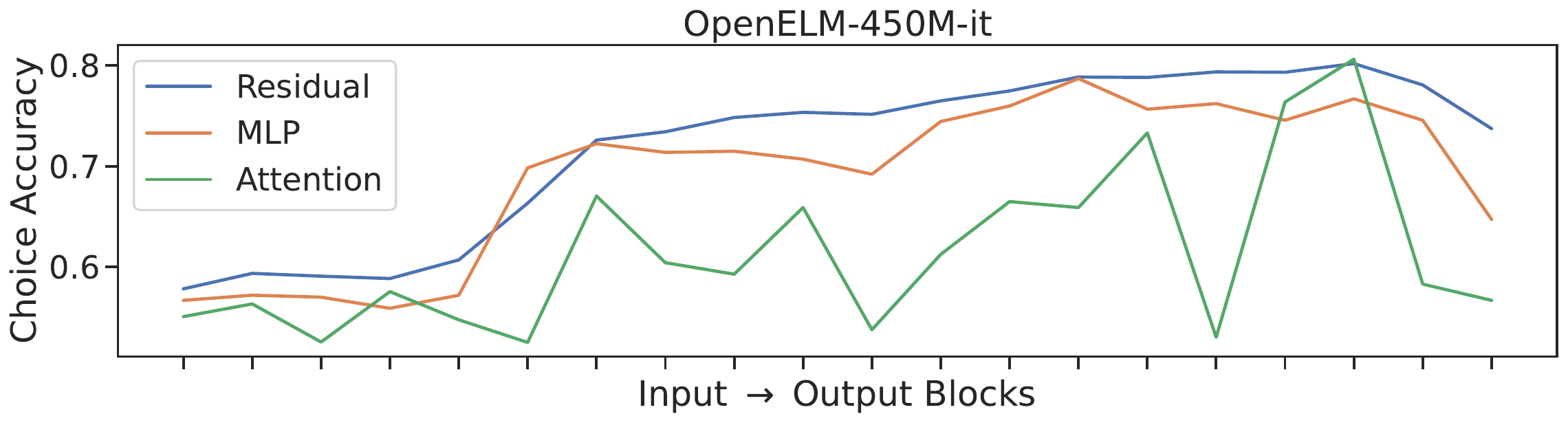}\par
        \includegraphics[width=\linewidth]{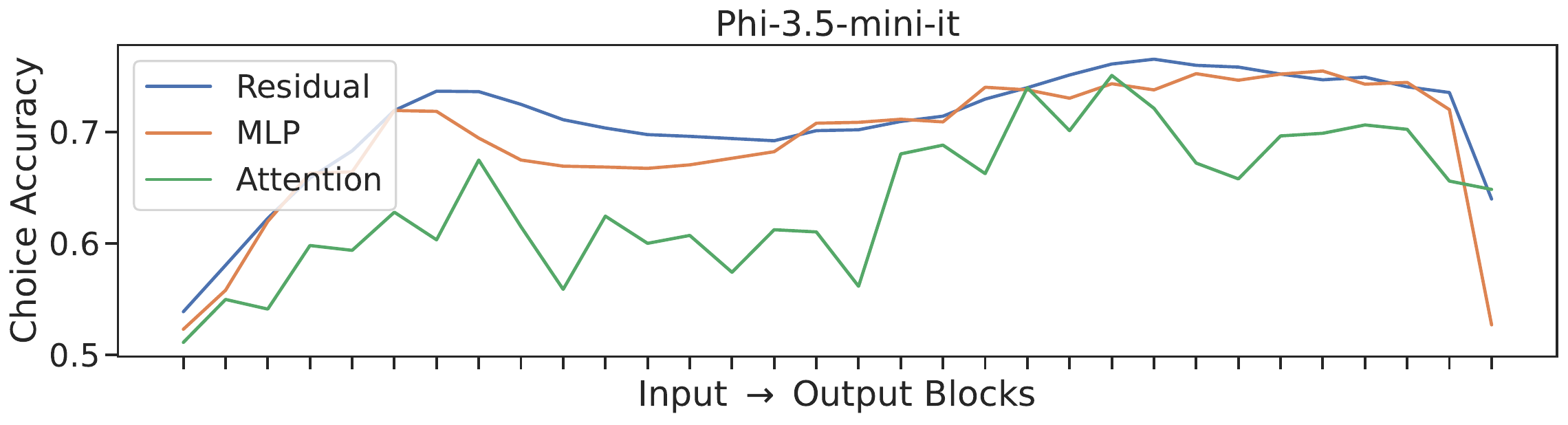}\par
        \includegraphics[width=\linewidth]{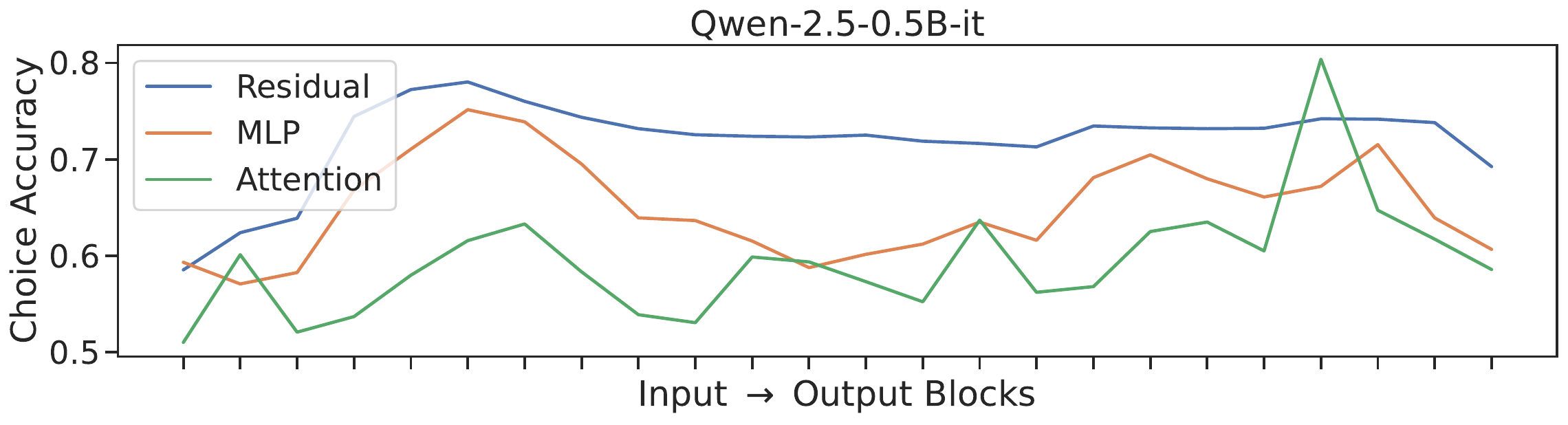}\par
        \includegraphics[width=\linewidth]{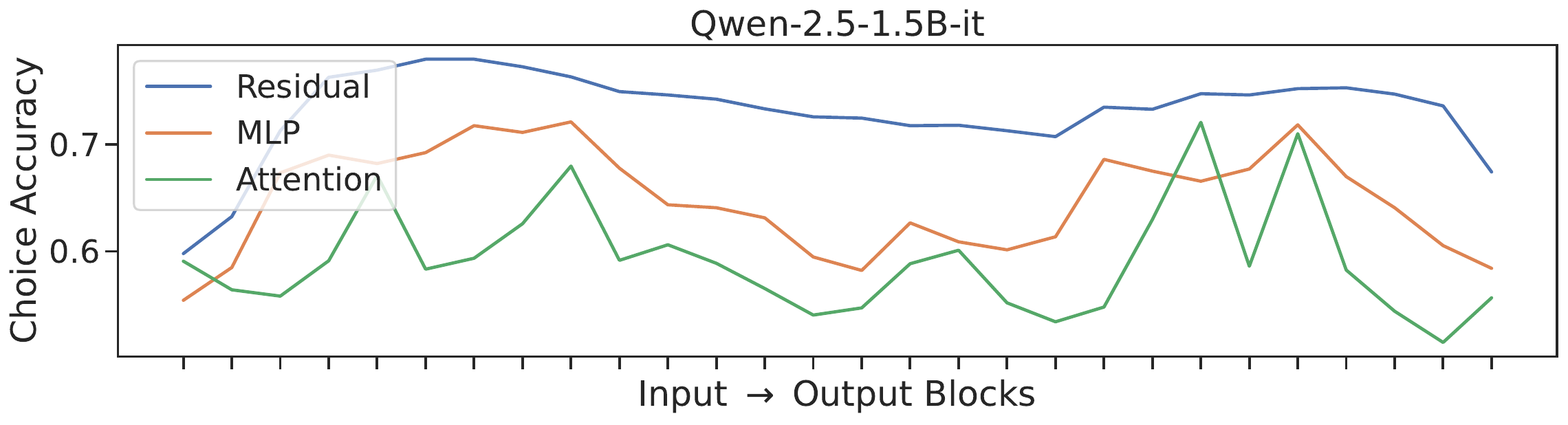}\par
        \includegraphics[width=\linewidth]{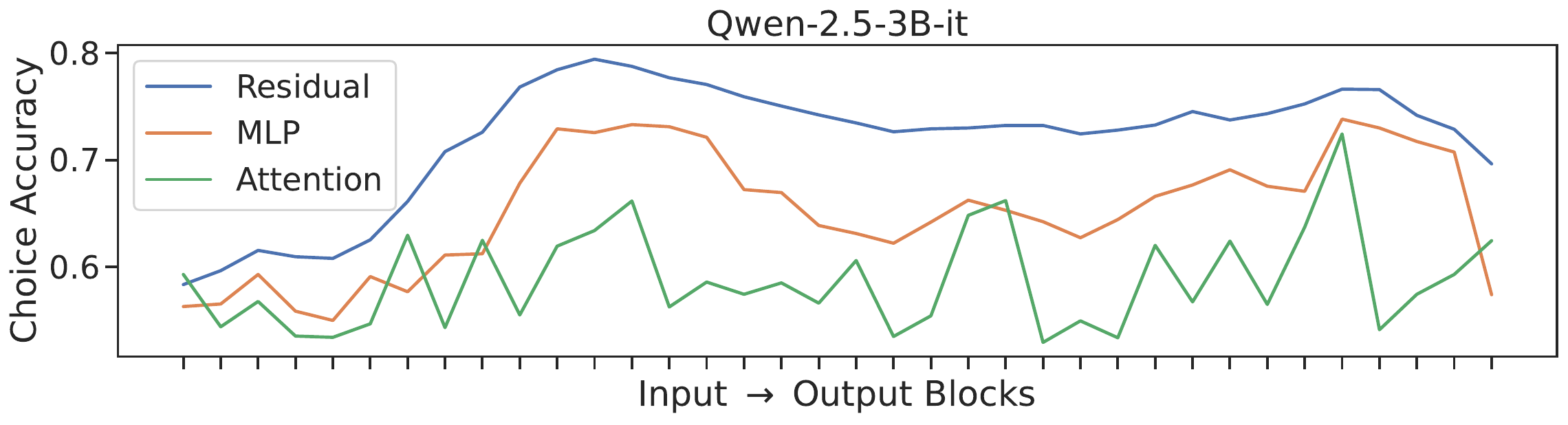}\par
        \includegraphics[width=\linewidth]{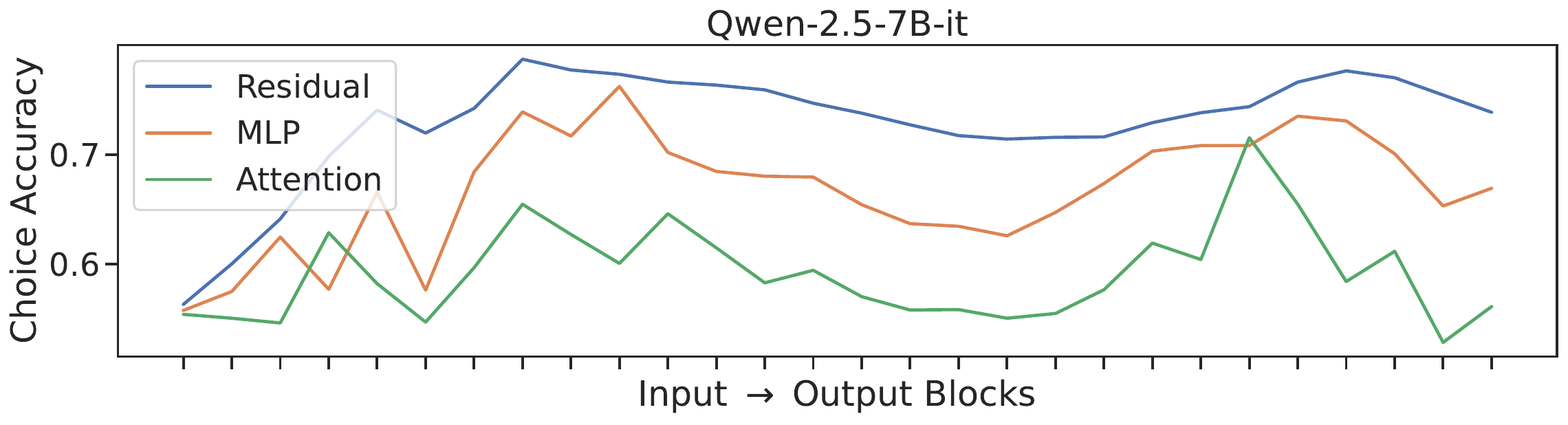}\par
    \end{multicols}
    \caption{Development of choice accuracy over layers for all instruction-tuned models.}
    \label{appx:fig:ca_all_it}
\end{figure}

In this section, we report the choice accuracy of the representations obtained from the attention layer, MLP layer, and residual stream of every block of every evaluated pretrained model. It can be seen in Fig.\ref{appx:fig:ca_all_pt} and Fig.\ref{appx:fig:ca_all_it} that qualitatively, the choice-accuracy dynamics over layers varies across models. 

To quantify the observation of relatively high smoothness of residual stream layers, we calculate the total variation across layers for every layer type separately:
\begin{align}
    \text{tv} =  \sum_{l=2}^L |\text{ca}_l - \text{ca}_{l-1}|
\end{align}
Here $\text{ca}_l$ is the choices accuracy at layer $l$ of $L$ layers of the same layer type. The aggregated results over all models in Fig.~\ref{appx:fig:tx_boxes} confirm that residual stream layers are the most consistent in their similarity structure, whereas attention layers are the most volatile.

\section{Additional Analysis on \texorpdfstring{$\gamma$}{gamma}}\label{appx:calibration}
    \begin{figure}[htb!]
        \subfloat{%
            \includegraphics[width=.49\linewidth]{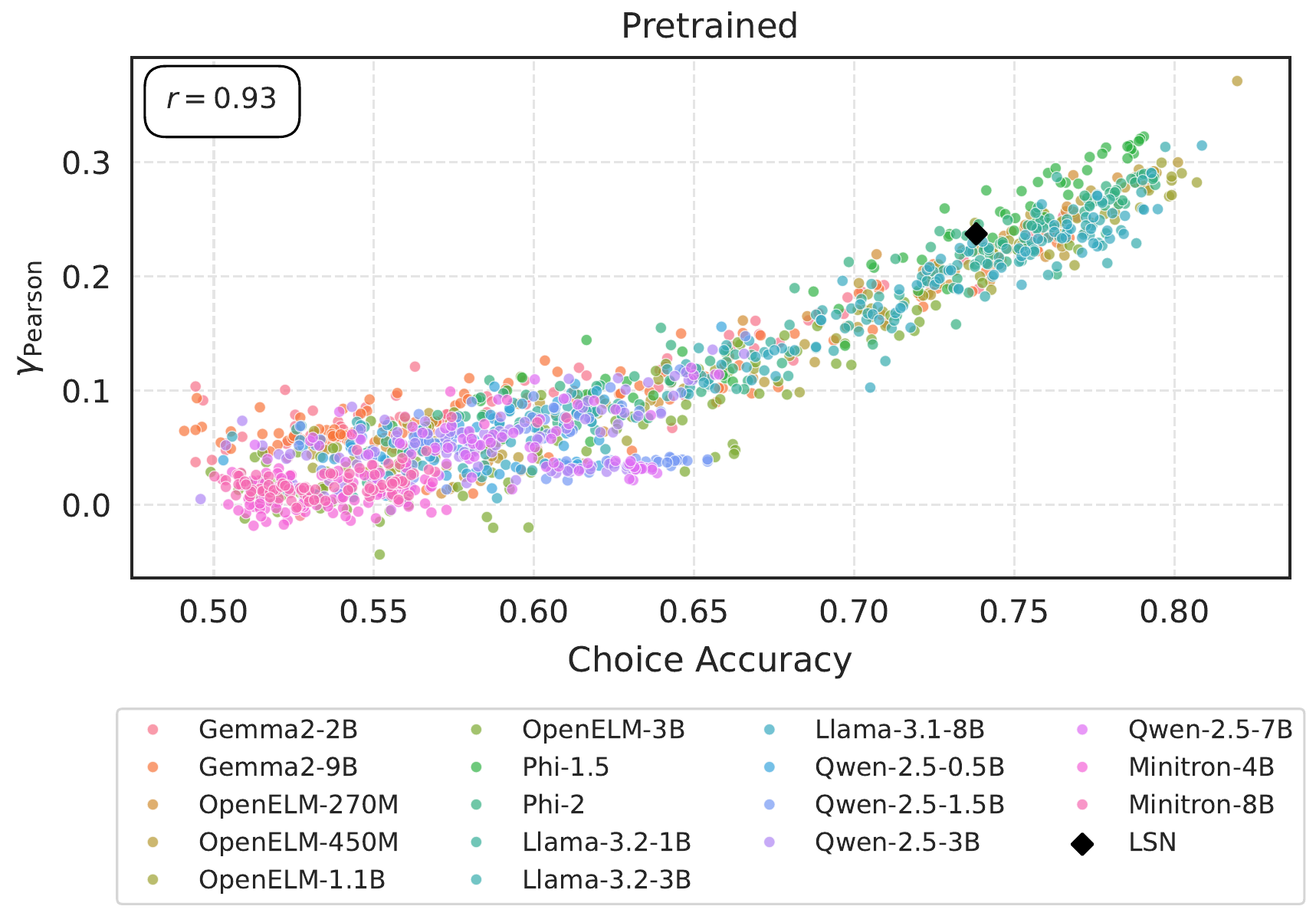}%
        }\hfill
        \subfloat{%
            \includegraphics[width=.49\linewidth]{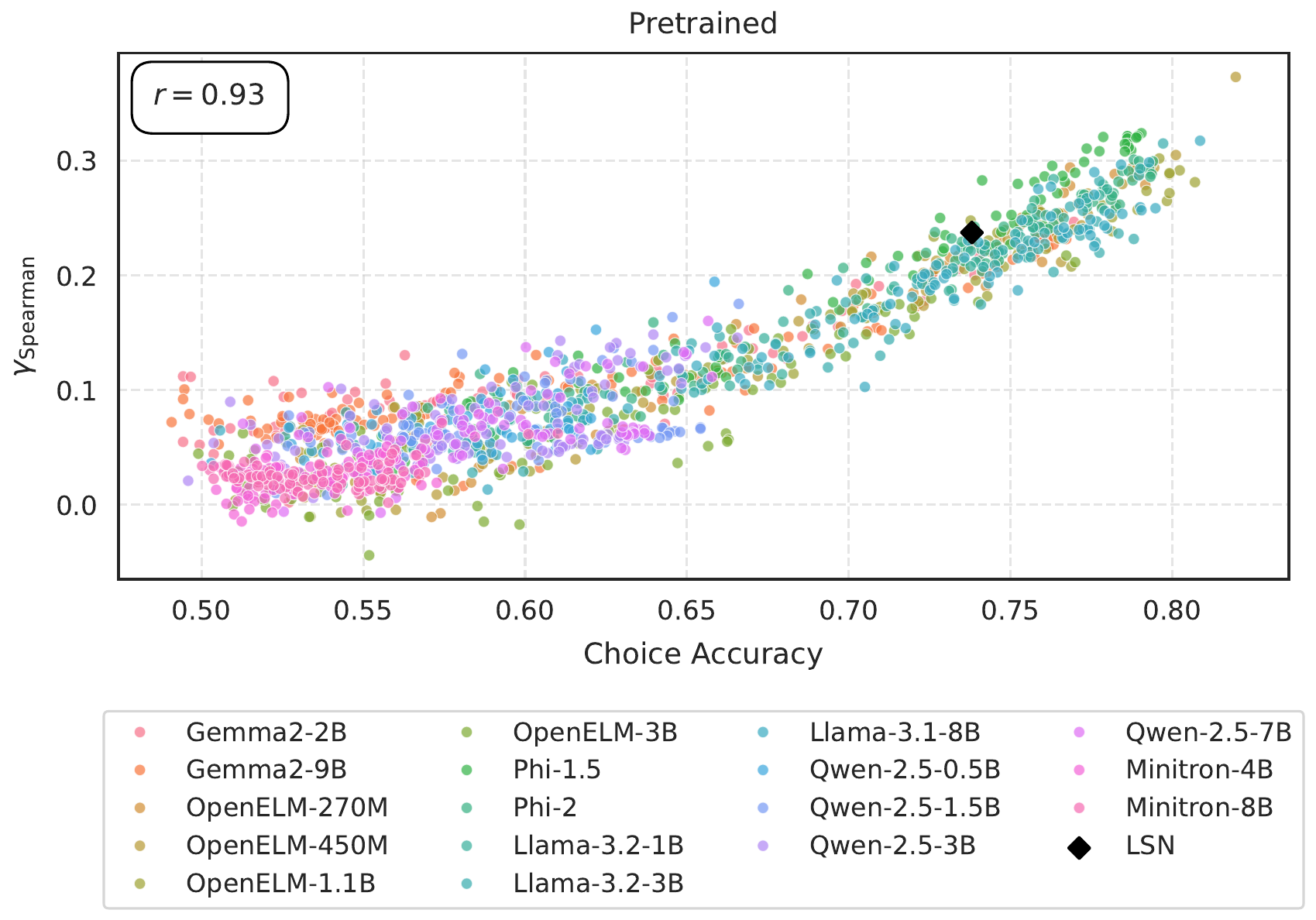}%
        }\\
        \subfloat{%
            \includegraphics[width=.49\linewidth]{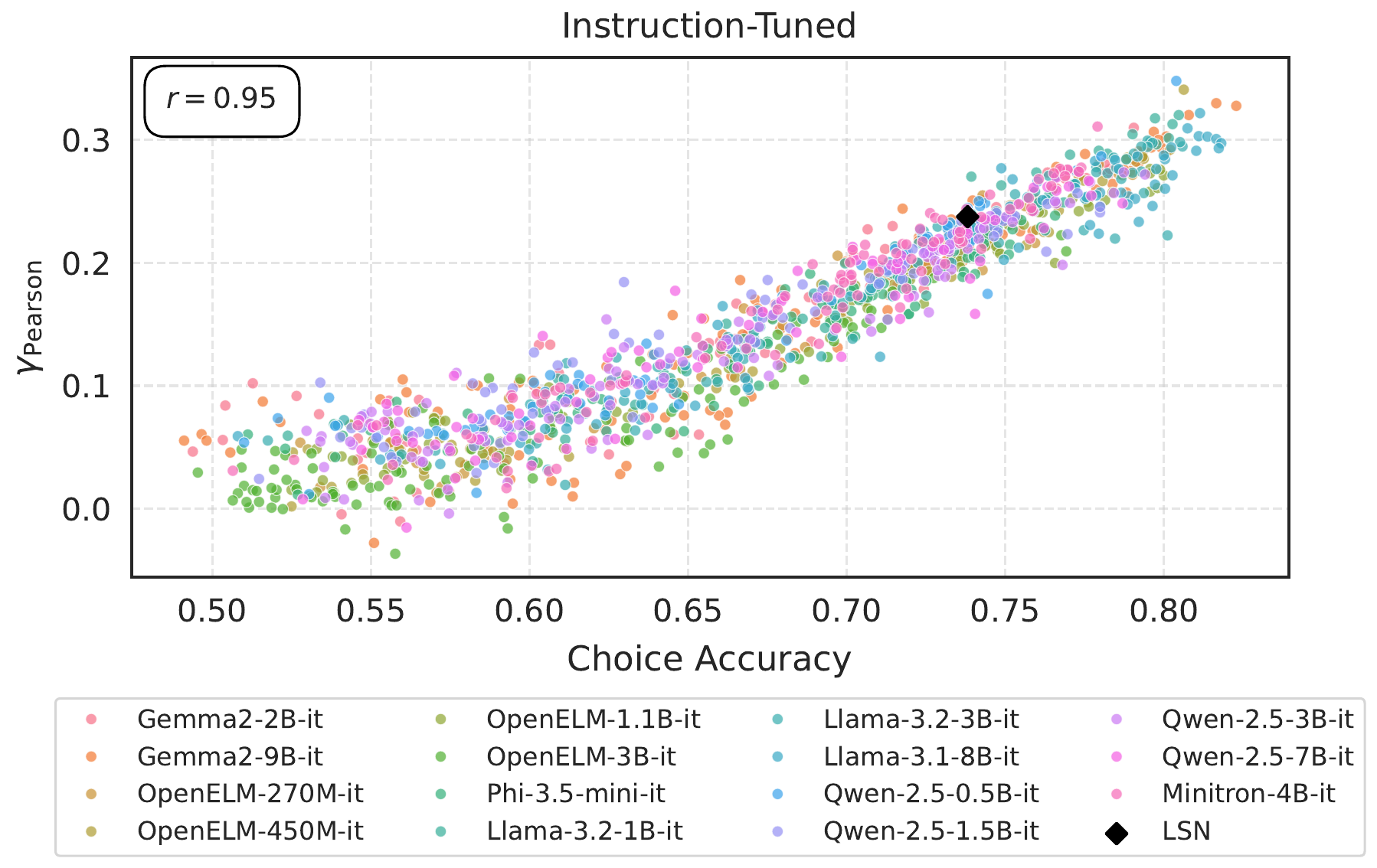}%
        }\hfill
        \subfloat{%
            \includegraphics[width=.49\linewidth]{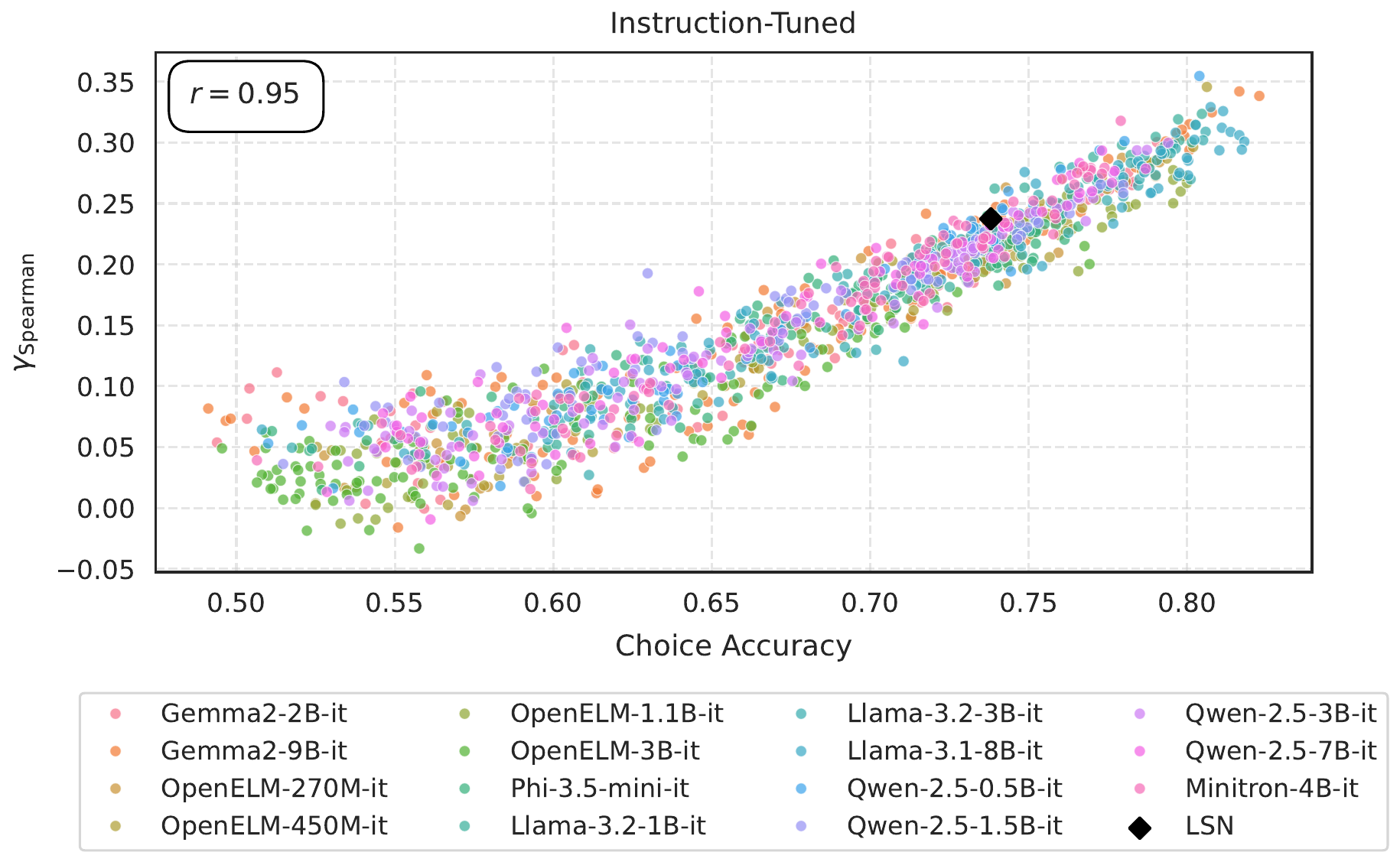}%
        }
        \caption{$\gamma$ v.s. choice-accuracy for pretrained (\textbf{top}) and instruction-tuned (\textbf{bottom}) models, as well as for computations of $\gamma$ based on Pearson (\textbf{left}) and Spearman (\textbf{right}) correlation coefficients. The r-value at the top left of each plot provides the Pearson correlation coefficient of the values on the x-axis and the y-axis.}
        \label{appx:fig:calibration-comparison}
    \end{figure}

In this section, we report results for the experiment on $\gamma$, correlating human agreement and $1-$ distance ratio, from Sec.~\ref{sec:ex:unc} of the main text. We use Spearman and Pearson correlation coefficients, as well as pretrained and instruction-tuned models. 

As can be seen in Fig.~\ref{appx:fig:calibration-comparison}, the trend that models showing a higher choice accuracy also achieve higher $\gamma$ is preserved when using Spearman correlation. For pretrained models, results are mostly comparable to the ones obtained on instruction-tuned models, with the only notable difference being a set of layers underperforming in $\gamma$, relative to their choice accuracy. This set is mainly comprised of the residual stream layer of Qwen and Minitron models -- the two model families seeing the largest improvement by instruction tuning.

\section{Examples of Triplets}\label{appx:examples}
\begin{table}[htb]
    \centering
    \setlength{\tabcolsep}{4pt}
\begin{subtable}[c]{0.45\textwidth}
\begin{tabular}{l|l|l}
\toprule
Anchor & Target 1 & Target 2 \\
\midrule
waterwheel & \textbf{watermill} & fingerlike \\
touchpad & \textbf{mousepad} & midfield \\
fruit & \textbf{pear} & merchandise \\
truck & \textbf{tanker} & upholstery \\
clay & \textbf{sandstone} & javelin \\
chocolate & \textbf{butterscotch} & tourist \\
stroller & \textbf{pushchair} & harbor \\
telescope & \textbf{binoculars} & university \\
lollipop & \textbf{lollypop} & officer \\
hand & \textbf{finger} & face \\
\bottomrule
\end{tabular}
\end{subtable}
\begin{subtable}[c]{0.45\textwidth}
\begin{tabular}{l|l|l}
\toprule
Anchor & Target 1 & Target 2 \\
\midrule
baseball & \textbf{outfielder} & cricket \\
jam & \textbf{blackberry} & jump \\
root & \textbf{woody} & proxy \\
surfboard & \textbf{pier} & toothbrush \\
curb & spur & \textbf{floorboard} \\
screw & \textbf{prosthetics} & buck \\
cat & rat & \textbf{meow} \\
dryer & \textbf{lint} & dishwasher \\
guacamole & margarita & \textbf{chip} \\
comb & \textbf{flyer} & wort \\
\bottomrule
\end{tabular}
\end{subtable}
    \caption{Examples of triplets with low and high human-model agreement for \textit{pretrained} models. (\textbf{Left}) all models chose the same target as the human majority, (\textbf{right}) all models chose the other target. Inter-human agreement is high in all examples. Terms in \textbf{bold} were chosen by the human majority vote.}
    \label{appx:tab:examples}
\end{table}

\begin{table}[htb]
    \centering
    \setlength{\tabcolsep}{4pt}
\begin{subtable}[c]{0.45\textwidth}
\begin{tabular}{l|l|l}
\toprule
Anchor & Target 1 & Target 2 \\
\midrule
waterwheel & \textbf{watermill} & fingerlike \\
fruit & \textbf{apple} & spice \\
door & \textbf{gate} & desk \\
money & \textbf{cash} & planetarium \\
pepperoni & \textbf{pizza} & scrubber \\
pocket & \textbf{bag} & panhandle \\
sailboat & \textbf{powerboat} & grandmother \\
prism & \textbf{refraction} & mentality \\
projector & \textbf{auditorium} & collie \\
scarf & \textbf{hairnet} & land \\
\bottomrule
\end{tabular}
\end{subtable}
\begin{subtable}[c]{0.45\textwidth}
\begin{tabular}{l|l|l}
\toprule
Anchor & Target 1 & Target 2 \\
\midrule
baseball & \textbf{outfielder} & cricket \\
jam & \textbf{blackberry} & jump \\
box & picture & \textbf{tupperware} \\
breakfast & \textbf{shaker} & gratitude \\
filter & speed & \textbf{rainwater} \\
train & training & \textbf{railroad} \\
goldfish & \textbf{pond} & hamster \\
surfboard & \textbf{pier} & toothbrush \\
screw & \textbf{prosthetics} & buck \\
hanger & hangar & \textbf{garter} \\
\bottomrule
\end{tabular}
\end{subtable}
    \caption{Examples of triplets with low and high human-model agreement for \textit{instruction-tuned} models. (\textbf{Left}) all models chose the same target as the human majority, (\textbf{right}) all models chose the other target. Inter-human agreement is high in all examples. Terms in \textbf{bold} were chosen by the human majority vote.}
    \label{appx:tab:examples_instr}
\end{table}

\begin{table}[htb]
    \centering
    \setlength{\tabcolsep}{4pt}
\begin{subtable}[c]{0.45\textwidth}
\begin{tabular}{l|l|l}
\toprule
Anchor & Target 1 & Target 2 \\
\midrule
robe & \textbf{turban} & coliseum \\
guardrail & \textbf{roadway} & jabbed \\
prism & \textbf{refraction} & mentality \\
surfboard & \textbf{paddle} & hairstylist \\
mustache & \textbf{beard} & catsup \\
mustache & \textbf{beard} & carburetor \\
jellyfish & passage & \textbf{plankton} \\
headband & \textbf{headpiece} & ritalin \\
crystal & gut & \textbf{sapphire} \\
dress & \textbf{apparel} & pee \\
\bottomrule
\end{tabular}
\end{subtable}
\begin{subtable}[c]{0.45\textwidth}
\begin{tabular}{l|l|l}
\toprule
Anchor & Target 1 & Target 2 \\
\midrule
thumbtack & \textbf{corkboard} & prong \\
avocado & \textbf{macadamia} & mayo \\
noodle & \textbf{popsicle} & teahouse \\
pigeon & \textbf{kingfisher} & courier \\
meatloaf & pantry & \textbf{pumpernickel} \\
kite & \textbf{paraglider} & string \\
swing & \textbf{paddle} & jazz \\
mixer & encoder & \textbf{sealer} \\
retainer & solicitor & \textbf{spacer} \\
bike & \textbf{backpack} & jockey \\
\bottomrule
\end{tabular}
\end{subtable}
    \caption{Examples of triplets with low and high human-model agreement for \textit{behavioral responses of instruction-tuned} models. (\textbf{Left}) all models chose the same target as the human majority, (\textbf{right}) all but one models chose the other target. Inter-human agreement is high in all examples. Terms in \textbf{bold} were chosen by the human majority vote.}
    \label{appx:tab:examples_behav}
\end{table}

Tab.~\ref{appx:tab:examples} shows examples of triplets where the maximum-choice-accuracy layers of all pretrained models consistently agree or disagree with the human majority. All of these triplets have been selected for high inter-human agreement and are sorted by this agreement. 

While we cannot draw strong conclusions from this limited set of examples alone, it seems possible that human similarity judgments are at times based on association (\textit{cat} and \textit{meow}, \textit{dryer} and \textit{lint}, \textit{surfboard} and \textit{pier}) whereas language models' similarity judgments are more often based on type (\textit{cat} and \textit{rat}, \textit{dryer} and \textit{dishwasher}, \textit{surfboard} and \textit{toothbrush}). Future work should include larger, quantitative evaluations to assess the difference in how humans and language models construct similarity judgments.

Tab.~\ref{appx:tab:examples_instr} and Tab.~\ref{appx:tab:examples_behav} show examples for representational and behavioral choices of instruction-tuned models. We note some overlap between the pretrained and instruction-tuned representational examples for which models disagree with the human majority.

\clearpage

\section{Results for Distilled DeepSeek-R1} \label{appx:deepseek}

In this section, we extend our basic choice accuracy evaluation to two distilled versions of DeepSeek-R1~\citep{deepseekai2025}, that use Qwen-Math-1.5B\footnote{\url{https://huggingface.co/deepseek-ai/DeepSeek-R1-Distill-Qwen-1.5B}} and Qwen-Math-7B\footnote{\url{https://huggingface.co/deepseek-ai/DeepSeek-R1-Distill-Qwen-7B}} as base models. 

For behavioral response extraction, we allowed the models to generate up to \num{2000} tokens, and then post-processed the output following the reasoning as detailed in Appx.~\ref{appx:method}. We observe that for some triplets, the models appear to get stuck in a reasoning loop without making a choice within the allotted token budget. We count these cases as invalid answers, resulting in an invalid answer fraction of 0.22 for the 1.5B model and 0.07 for the 7B model. We note that this fraction may be reduced by tuning the temperature parameter.

\begin{figure}[htb!]
    \centering
        \includegraphics[width=\linewidth]{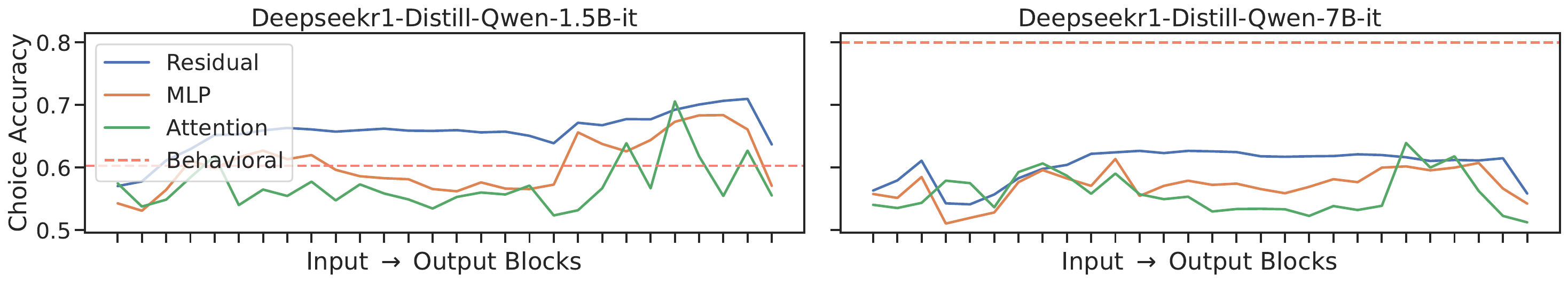}
    \caption{Choice accuracy across layers for two distilled DeepSeek-R1 variants.}
    \label{appx:fig:deepseek-layer-ca}
\end{figure}

In Fig.~\ref{appx:fig:deepseek-layer-ca}, it can be seen that the representational choice accuracy is comparably low for both models (0.64 and 0.71). The 8B variant achieves a significantly higher behavioral choice accuracy of 0.80. We speculate that this discrepancy of representational and behavioral performance may stem from a reduced need for easily decodable representations, as this decoding can be done by the model over lengthy reasoning chains. We note that further analyses are warranted, as in these two models, the reasoning component is confounded with the math-focused training of the base model.

\section{Prompt Robustness Analysis} \label{appx:promt-robust}

\begin{figure}[htb!]
    \centering
        \includegraphics[width=0.49\linewidth]{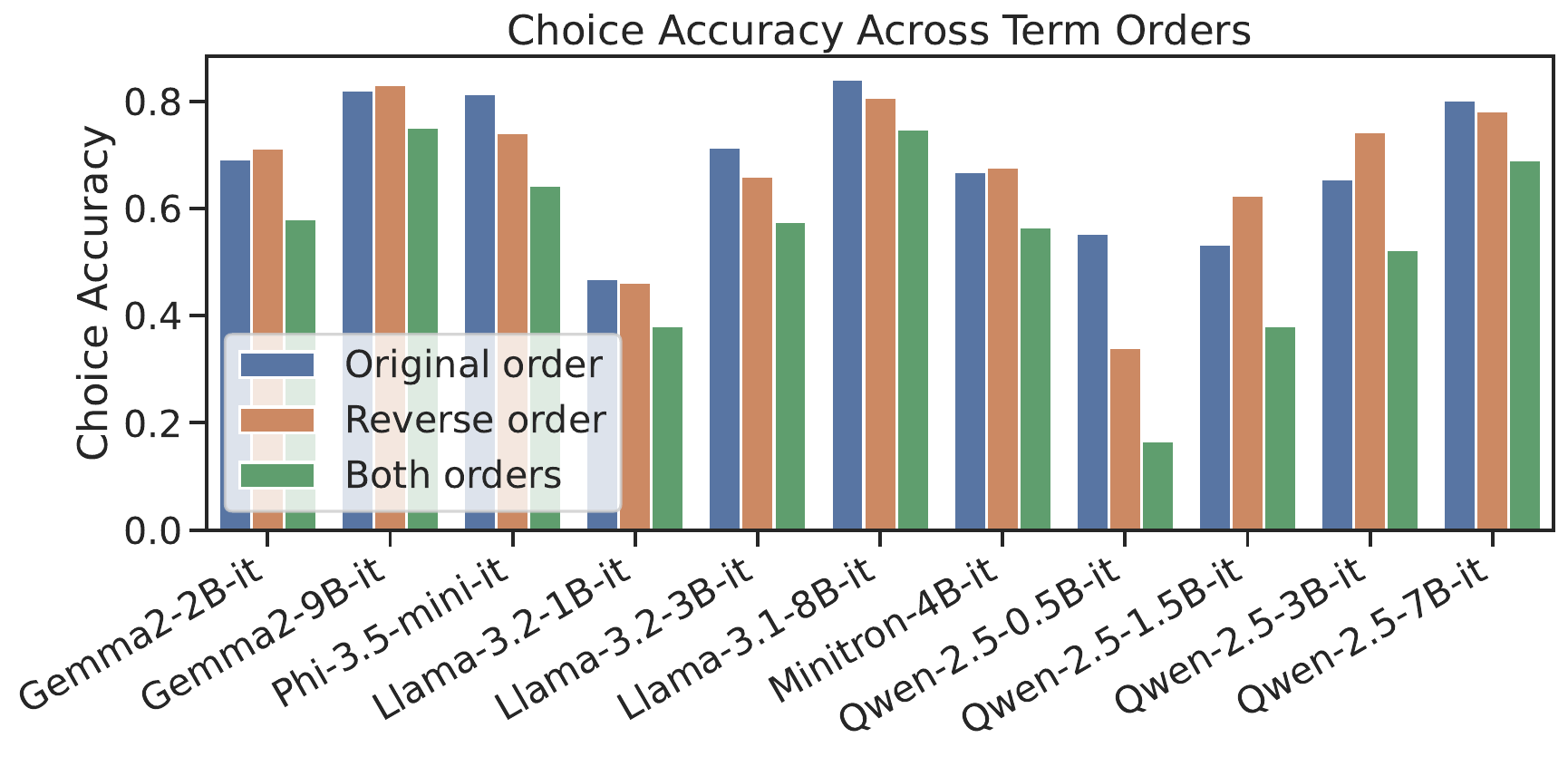}
        \includegraphics[width=0.49\linewidth]{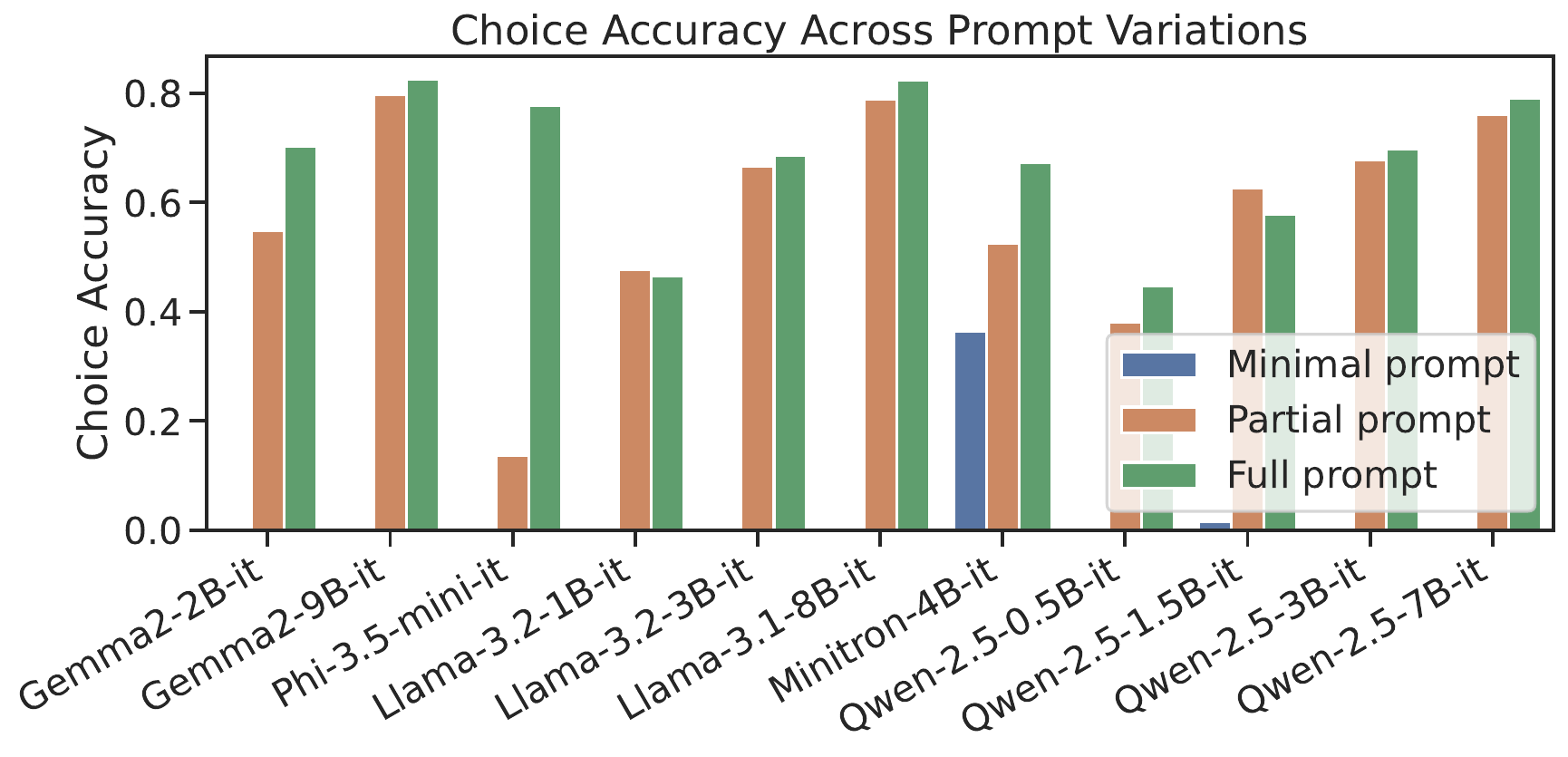}
    \caption{(\textbf{Left}) Behavioral choice accuracy for the original target term order in the 3TT datasets, reversed term order, and when only counting triplets as correct if they have been answered correctly in both orders. (\textbf{Right}) Behavioral choice accuracy for three variations of the prompt. Only instruction-tuned models are shown. OpenELM models are excluded due to poor instruction following. }
    \label{appx:fig:prompt-order}
\end{figure}

To evaluate the robustness of our results w.r.t prompt design, we provide two additional experiments investigating (1) the effect of the order in which the two target terms are presented within the prompt, and (2) the effect of varying the prompt formulation.

In Fig.~\ref{appx:fig:prompt-order}~(left), it can be seen that the order in which the target terms are presented can impact choice accuracy. Yet, this impact is only large for the smaller Qwen models. It should be noted that due to the dataset construction, the first term, according to the original order, has a higher chance of being more similar to the anchor. This results in the human choice being the first term in 70\% of the evaluated triplets. This explains why term order can influence choice accuracy and supports our strategy of evaluating both orders.

In Fig.~\ref{appx:fig:prompt-order}~(right) we compare three variations of the prompt: The \textbf{minimal} prompt reads \textit{``Which of the words A or B is closer in meaning with the word C?"}, the \textbf{partial} prompt additionally ends with \textit{``Answer with exactly one word: either A or B."}, and the \textbf{full} prompt adds to the partial prompt \textit{``Do not answer with C. Do not answer in a full sentence."} The minimal prompt was used for gathering the human responses in the 3TT dataset, whereas the full prompt was used in the main part of this paper. It can be seen that the relative ordering of the models' choice accuracies does not greatly change across the latter two variations, indicating some robustness to the prompt formulation. Here, the exception is Phi-3.5-mini, which only performs well on the full prompt. Overall, the shorter variations of the prompt show reduced choice accuracy (around 0 for most models in the minimal formulation). We attribute this to answers being incompatible with our answer evaluation schema (see Appx.~\ref{appx:method}), and then being counted as invalid answers.

\end{document}